\documentclass{article}

\usepackage{hyperref}
\usepackage{booktabs}
\usepackage{graphics}
\usepackage[table,xcdraw,dvipsnames]{xcolor}
\usepackage{color}
\usepackage{listings}
\usepackage{natbib}
\usepackage{graphicx}
\usepackage{iclr2021_conference,times} % Use this for the initial submission.

\usepackage[font=small,labelfont=bf]{caption}
\usepackage{wrapfig}
\usepackage[font=scriptsize]{subcaption}

\newcommand{\reviewer}[3]{
	\expandafter\newcommand\csname #1\endcsname[1]{
		\textcolor{#3}{#2: ##1}
	}
}

\usepackage{xspace}
\newcommand{\algoName}{\texttt{D2RL}\xspace}
\usepackage{enumitem}

\usepackage{titlesec}
\titlespacing*{\section}
{0pt}{.4\baselineskip}{0.5\baselineskip}
\titlespacing*{\subsection}
  {0pt}{0.25\baselineskip}{0.2\baselineskip}

\definecolor{codegreen}{rgb}{0,0.6,0}
\definecolor{codegray}{rgb}{0.5,0.5,0.5}
\definecolor{codepurple}{rgb}{0, 0, 0}
\definecolor{backcolour}{rgb}{0.95,0.95,0.92}

\lstdefinestyle{mystyle}{
    backgroundcolor=\color{backcolour},   
    commentstyle=\color{magenta},
    keywordstyle=\color{magenta},
    numberstyle=\tiny\color{codegray},
    stringstyle=\color{codepurple},
    basicstyle=\ttfamily\footnotesize,
    breakatwhitespace=false,         
    breaklines=true,                 
    captionpos=b,                    
    keepspaces=true,                 
    numbers=left,                    
    numbersep=5pt,                  
    showspaces=false,                
    showstringspaces=false,
    showtabs=false,                  
    tabsize=2,
    language=Python,
}

\reviewer{aravind}{AS}{black}
\reviewer{Homanga}{HB}{red}
\reviewer{Sam}{Sam}{green}
\reviewer{Animesh}{AG}{magenta}

\title{D2RL: Deep Dense Architectures in Reinforcement Learning}

\author{Samarth Sinha$^{1*}$, Homanga Bharadhwaj$^{1}$\thanks{equal contribution}, Aravind Srinivas$^2$, \textbf{Animesh Garg}$^1$\\
$^1$University of Toronto, Vector Institute\\
$^2$University of California Berkeley\\
\texttt{samarth.sinha@mail.utoronto.ca}, \texttt{homanga@cs.toronto.edu} 
}

\iclrfinalcopy 
\begin{document}\maketitle

%===============================================================================

\begin{abstract}
% Incorporating appropriate inductive biases in deep reinforcement learning (DRL) has the potential to improve sample efficiency, making typical reinforcement learning pipelines requiring millions of environment interactions efficient enough to be deployed for controlling real robots. Although recent works have introduced inductive biases of invariance, when learning from images, through techniques such as data augmentations and contrastive losses, these efforts have mostly focused on representation learning, while largely ignoring the investigation of architectures for policy and value functions. In this paper, we first perform an empirical study of increasing the depth of typical feedforward neural networks used in DRL and show how performance drops with increasing depth. Since deeper layers help in extracting richer features, we draw inspiration from computer vision and generative modeling literature to design a modified architecture based on skip-connections. Across a suite of various DRL algorithms, in a set of robotic control environments involving locomotion and manipulation with different robots, we demonstrate significant performance gains with the proposed architectural variant applied to the parameterization of the policy and value functions.
% We present an empirical investigation of deep learning architectures for continuous control benchmarks. 
While improvements in deep learning architectures have played a crucial role in improving the state of supervised and unsupervised learning in computer vision and natural language processing, neural network architecture choices for reinforcement learning remain relatively under-explored. 
We take inspiration from successful architectural choices in computer vision and generative modeling, and investigate the use of deeper networks and dense connections for reinforcement learning on a variety of simulated robotic learning benchmark environments. 
Our findings reveal that current methods benefit significantly from dense connections and deeper networks, across a suite of manipulation and locomotion tasks, for both proprioceptive and image-based observations. We hope that our results can serve as a strong baseline and further motivate future research into neural network architectures for reinforcement learning. The project website is at this link \url{https://sites.google.com/view/d2rl/home}%We have released code for our experiments in the project website \href{https://sites.google.com/view/d2rl/home}{https://sites.google.com/view/d2rl/home} .
% Research Question:\\
% Hypothesis: U-net type policy arch works better than feedforward fully connected arch (do we compare against recurrent architectures?)\\
% Intuition: Why do we think our hypothesis is valid?\\
% Findings:
\end{abstract}

% \keywords{Inductive Bias, Deep Reinforcement Learning, Neural Network Architectures for Robot Learning} 

%===============================================================================

\section{Introduction}

Deep Reinforcement Learning (DRL) is a general purpose framework for training goal-directed agents in high dimensional state and action spaces. There have been plenty of successes from DRL for robotic control tasks, spanning across locomotion and navigation tasks, both in simulation and in the real world \citep{trpo, rubiks, qtopt}.

While the generality of the DRL framework lends itself to be applicable to a wide variety of tasks, one has to address issues such as the sample-efficiency and generalization of the agents trained with this framework. Sample-efficiency is fundamentally critical to agents trained in the real world, particularly for robotic control tasks. Baking in {\it minimal} inductive biases into the framework is one effective mechanism to address the issue of sample-efficiency of DRL agents and make them more efficient. 

%Deep Reinforcement Learning (DRL) has been very successful in robotic control tasks~\citep{qtopt,trpo,rubiks} due to the generality of the framework and its efficacy in requiring minimal supervision.
%The generality of the DRL framework also has its disadvantages in being unsafe, less optimal for particular tasks compared to imitation learning based solutions. 
The generality of the framework makes it difficult to control particular behaviours and inductive biases for DRL algorithms.
%To this end, it is important to include relevant inductive biases in the learning process, as pointed out by multiple recent works~\citep{}. 
Inductive biases are important for learning algorithms, as they are able to induce desirable behaviour in the learned agents.
Recent work has sought to improve the sample efficiency of DRL by adding an inductive bias of invariance, when learning from images, through techniques such as data augmentations~\citep{rad,yarats} and contrastive losses~\citep{curl}.
Similarly, another important inductive bias in DRL is the choice of the architectures for function approximators, for example how to parameterize the neural network for the policy and value functions. 
However, the problem of choosing architecture designs in DRL and robotics, for planning and control, has been largely ignored.

%There has been very less effort devoted to the study of suitable architectures for the value and policy functions in DRL. 
%Most current DRL papers still keep using 84x84 grayscale images [cite], a legacy carried on from the original DQN papers~\citep{}. In addition, effects of normalization layers like BN, LN, GN~\citep{} are underexplored in DRL compared to the understanding we have in computer vision, natural language processing and generative modeling. Trying to incorporate meaningful inductive biases in learning better parameterizations of the function approximators is important as it might lead to faster learning. It is important to be able to learn effective policies using as few environment interactions as possible when deploying DRL to real robots. Algorithms that require millions of environment interactions while learning in robotic simulations and games operating in faster than real time will be very slow and inefficient when deployed in real robots that operate in real time. In Computer Vision, we observed similar speedups in learning after the introduction of simple inductive biases like residual connections and BatchNorm, especially when deploying deep learning models on large datasets like ImageNet~\citep{}.  

Modern computer vision and language processing research have shown the disproportionate advantage of the size and depth of the neural networks used~\citep{resnet1001, radford2019language} wherein \textit{very} deep neural networks can be trained such that they learn better and more generalizable representations. Furthermore, recent evidence suggests that deeper neural networks can not only learn more complex functions but also have a smoother loss landscape~\citep{davidsdeepnetworkspaper}. Learning function approximators which enable better optimization and expressivity is an important inductive bias, which is greatly exploited in vision and language processing by using clever neural network architecture choices such as residual connections~\citep{resnet}, normalization layers~\citep{bn}, and gating mechanisms~\citep{lstm}, to name a few. It would be ideal to incorporate similar inductive biases in modern DRL algorithms in robotics in order to allow for better sample efficiency as that would significantly aid the deployment of real world robot learning agents.

% In this paper, we first highlight the problems that occur when learning policies and value functions using deep-er neural networks.
% We then propose a simple yet effective way to alleviate the problem such that we are able to utilize the benefits of parameterizing policies and value functions using deeper networks in a variety of DRL and robotics settings.
% We observe that by
%we focus on method which is to incorporate inductive biases in DRL via the parameterization of the learned models. 
%We postulate that changes with respect to model designs are even more general than data in robot learning, as sometimes it is unclear how to perform data augmentations in tasks where a robot is say tasked to manipulate objects in a pseudo-static environment. 

% incorporate meaningful inductive biases in the learning process of DRL from pixel inputs, so as to improve performance in solving complex tasks. Often, there is a tradeoff between including more inductive biases to be good at particular tasks and including less inductive biases to be (less) good at a wider range of tasks. Most previous approaches of learning control policies directly from image inputs have sought to improve the representation learning phase, by incorporating contrastive learning loss as in CURL~\citep{curl}, data augmentations as in RAD~\citep{rad} and Yarats~\citep{yarats}, relational inductive biases~\citep{}, and learning higher dimensional latent representations~\citep{}

In this paper, we first highlight the problems that occur when learning policies and value functions using vanilla deep neural networks.
Then we propose \algoName; an architecture that addresses these problems while benefiting from the utility of inductive biases added by more expressive function approximators.
We show how our proposed architecture scales effectively to a wide variety of off-policy RL algorithms, for proprioceptive-feature and image based inputs across a diverse set of challenging robotic control and manipulation environments. 
Our approach is motivated by utilizing a form of dense-connections similar to the ones found in modern deep learning, such as DenseNet~\citep{densenet}, Skip-VAE~\citep{skipvae} and U-Nets~\citep{unet}.
%which are all powering plenty of commercial applications. 
%To the best of our knowledge, we are performing the first exploration of such architectures in the context of DRL for robotic control. 
We demonstrate that the proposed parameterization significantly improves sample efficiency of RL agents (as measured by the number of environment interactions required to obtain a level of performance) in continuous control tasks. %which is very important for deploying these algorithms to real robots that interact with the environment in real time. 
% As a consequence of our simplicity and the benefit from using hardware accelerators such as GPUs, %In addition to significantly improving sample efficiency, 
% we also demonstrate improvements with respect to wall clock training time of algorithms. 

Our contributions can be summarized as:

\begin{enumerate}[
    topsep=0pt,
    noitemsep,
    % partopsep=1ex,
    % parsep=1ex,
    leftmargin=*,
    % itemindent=3ex
    ]
    \item We investigate the problem with increasing the number of layers used to parameterize policies and value functions.
    \item We propose a general solution based on dense-connections to overcome the problem.
    \item We extensively evaluate our proposed architecture on a diverse set of robotics tasks from proprioceptive features and images across multiple standard algorithms.
\end{enumerate}

% \begin{itemize}
%     \item Inductive biases are important in ML. Why?
%     \item Important in RL. Why? Helps increase sample efficiency by exploiting domain knowledge of the problem/environment or by introducing some biases in the structure of the learning algorithm.
%     \item We don't target the use of domain knowledge in learning - that is a separate and interesting problem on its own.
%     \item What happens when the complexity class of the policy networks is arbitrarily increased? (e.g. by having deeper neural networks) Lessons from papers like ResNet...
%     \item Are some policy parametrerizations better than others? Intuition for why?
    
% \end{itemize}

\section{Related work}
% \begin{figure}
%   \centering
%   \hspace*{-0.3cm}
%       \begin{minipage}[c]{0.75\textwidth}
%      \includegraphics[width=0.9\textwidth]{plots/mainfig_modified_corl.png} 
%       \end{minipage}
%               \begin{minipage}[c]{0.25\textwidth}
%     \caption{Visual illustrations of the proposed dense-connections based \algoName modification to the policy $\pi_\phi(\cdot)$ and Q-value $Q_\theta(\cdot)$ neural network architectures. 
%     We do note that the dense connections in our proposed architectures are concatenation based and not additive such as in ResNets~\citep{resnet}.
%     The inputs are passed to each layer of the neural network through identity mappings. Forward pass corresponds to moving from left to right in the figure. For state-based envs, $s_t$ is the observed simulator state and there is no convolutional encoder.}
%     \label{fig:teaser}
%       \end{minipage}
%      \vspace*{-0.8cm}
%   \end{figure}
  
\begin{figure}
   \centering
     \includegraphics[width=0.95\textwidth]{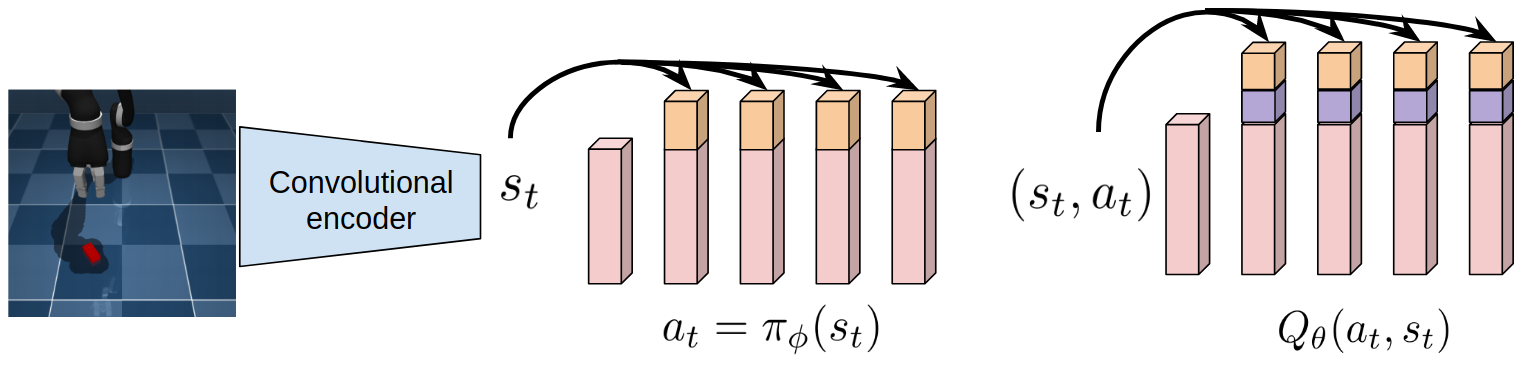} 
    \caption{Visual illustrations of the proposed dense-connections based \algoName modification to the policy $\pi_\phi(\cdot)$ and Q-value $Q_\theta(\cdot)$ neural network architectures. 
    % We do note that the residual connections in our proposed architectures are concatenation based and not additive such as in ResNets~\citep{resnet}.
    The inputs are passed to each layer of the neural network through identity mappings. Forward pass corresponds to moving from left to right in the figure. For state-based envs, $s_t$ is the observed simulator state and there is no convolutional encoder.}
    %\Animesh{use a better environment image either improve constrast or use a different env which is clean with perhaps white background. this is dark and nothing is discernible!}} \HB{done}
    \label{fig:teaser}
    \vspace*{-0.5cm}
  \end{figure}

% \begin{figure}[t]
%     \centering
%     \includegraphics[width=0.9\textwidth]{corl_architecture.png}
%     \caption{Policy Network (to be updated)}
%     \label{fig:mainfig}
% \end{figure}

\paragraph{Learning efficient representations for sample-efficient RL} 
Several recent papers have sought to improve representation learning of observations for control. CURL~\citep{curl} augments the usual RL loss with a contrastive loss that seeks to learn a latent representation which enforces the inductive bias of encodings of augmentations of the same image being closer in latent space than embeddings of augmentations of different images. RAD~\citep{rad} and DrQ~\citep{yarats}  showed that simple data augmentations like random crop, color jitter, patch cutout, and random convolutions can alone help improve sample-efficiency and generalization of RL from pixel inputs. Some other algorithms learn latent representations decoupled from policy learning, through a variational autoencoder based loss~\citep{higgins2017darla, dreamer, rig}. OFENet~\citep{inputdim} shows that learning a higher dimensional feature space helps learn a more informative representation when learning from states. SAC-AE~\citep{sac_ae} showed the utility of wide MLP layers (1024) in the actor and critic networks for SAC~\citep{sac} in pixel-based RL (Appendix B1 of~\citet{sac_ae}) . This architecture later propagated to CURL, RAD, and DrQ.~\citet{song2019observational} analyzed overfitting in model-free RL and found that using wider neural net architectures for the policy helps improve generalization.

\paragraph{Inductive biases in deep learning} 
Inductive biases in deep learning have been long explored in different contexts such as temporal relations~\citep{lstm}, spatial relations~\citep{cnn, krizhevsky2012imagenet}, translation invariance~\citep{mixmatch, moco, chen2020simple, curl, rad} and learning contextual representations~\citep{attention}.
These inductive biases are injected either directly through the network parameterization~\citep{cnn, lstm, attention} or by implicitly changing the objective function~\citep{mixmatch, curl}.

\paragraph{Learning very deep networks}
Deep neural networks are useful for extracting features from data relevant for various downstream tasks. 
However, simply increasing the depth of a feed-forward neural network leads to instability in training due to issues such as vanishing gradients~\citep{lstm}, or a a loss of mutual information~\citep{resnet}. 
To mitigate this, residual connections were proposed which involve an alternate path between layers through an identity mapping~\citep{resnet}.
Skip-VAEs~\citep{skipvae} tackle a similar issue of posterior collapse in typical VAE training by adding skip connections in the architecture of the VAE decoder. 
U-Nets~\citep{unet} consider a contractive path of convolutions and maxpooling layers followed by an expansive path of up-convolution layers. There are copy and crop identity mapping from layers of the contractive path to layers in the expansive path.
Normalization techniques such as batch normalization are also important in learning deep networks~\citep{batchnorm}.
Combining residual connections with batch normalization have been used to successfully train networks with $1000$ layers~\citep{resnet1001}.
Our proposed architecture closely resembles DenseNet~\citep{densenet}, which uses skip connections from feature maps of previous layers through concatenation, allowing for efficient learning and inference.

% we need to highlight the difference against https://arxiv.org/abs/2003.01629 (see fig 2) which deals with more of a representation learning aspect, but propose something quite similar

% What are some related papers for this? https://openreview.net/pdf?id=B1lqDertwr https://arxiv.org/pdf/2005.12729.pdf ?

% maybe some papers on inductive biases in RL

% https://arxiv.org/abs/1907.02908

% arXiv.orgarXiv.org

% On Inductive Biases in Deep Reinforcement Learning
% Many deep reinforcement learning algorithms contain inductive biases that sculpt the agent's objective and its interface to the environment. These inductive biases can take many forms, including...

% https://openreview.net/pdf?id=HkxaFoC9KQ

% ResNet for showing that depth alone does not help https://arxiv.org/abs/1512.03385

% arXiv.orgarXiv.org
% Deep Residual Learning for Image Recognition
% Deeper neural networks are more difficult to train. We present a residual learning framework to ease the training of networks that are substantially deeper than those used previously. We...

% This workshop might have some interesting content too https://sites.google.com/view/mibrl/home

% The UNet paper of course https://lmb.informatik.uni-freiburg.de/people/ronneber/u-net/

% Adji https://adjidieng.github.io/Papers/skipvae.pdf (closest architecture to ours)

\section{Preliminaries}

In this section, we describe the actor-critic formulation of RL algorithms, which serves as the basic framework for our setup. We then describe the Data-Processing Inequality which is relevant for explaining and motivating the proposed architecture. 

\subsection{Actor-Critic methods}
Actor-critic methods learn two models, one for the policy function and the other for the value function such that the value function assists in the learning of the policy. These are TD-learning~\citep{tdlearning} methods that have an explicit representation for the policy independent of the value function. We consider the setting where the critic is an state-action value function $Q_\theta(a,s)$ parameterized by a neural network, and the actor $\pi_\phi(a|s)$ is also parameterized by a neural network. Let the current state be $s$ and $r_t$ denote the reward obtained after executing action $a$ in state $s$, and transitioning to state $s'$. After sampling action $a'\sim\pi_\phi(a'|s)$ in the next state $s'$, the policy parameters are updated in the direction suggested by the critic $Q_\theta(a,s)$
$$\phi\leftarrow \phi + \beta_\phi Q_\theta(a,s)\nabla_\phi\log\pi_\phi(a|s) .$$

The parameters $\theta$ are updated using the TD correction $\Delta_t = r_t + \gamma Q_\theta(s',a') - Q_\theta(s,a)$ as follows: 
$$\theta \leftarrow \theta + \beta_\theta\Delta_t\nabla_\theta Q_\theta(s,a) .$$

Although the basic formulation above requires on-policy samples $(s,a,s',r)$ for gradient updates, a number of off-policy variants of the algorithm have been proposed~\citep{sac,td3} that incorporate importance weighting in the policy's gradient update. Let the observed samples be sampled by a behavior policy $a\sim\zeta(a|s)$, and $\pi_\phi(a|s)$ be the policy to be optimized. The gradient update rule changes as
$$\phi\leftarrow \phi + \beta_\phi\frac{\pi_\phi(a|s)}{\zeta(a|s)} Q_\theta(a,s)\nabla_\phi\log\pi_\phi(a|s) $$

% \subsection{Skip-connections in Neural Networks}
% Skip connections have been widely used to learn deep neural networks. 
% The network architecture of U-Net consists of a contractive path on the left and an expansive path on the right. The architecture is motivated by making efficient use of the available training samples through data augmentations. The contractive path is meant to capture context, while the expansive path is meant to enable effective localization. The architecture is symmetric, with copy operations directly mapping from layers on the left to layers on the right.

% The contractive path is based on the architecture of a typical CNN, and involves repeated application of two 3x3 convolutions and 2x2 max pooling operations. Each 3x3 convolution is followed by ReLU non-linearities, and the max pooling operations have a stride of 2. Reverse operations are present in the expansive path on the right. Instead of 2x2 max pooling layers, there are 2x2 up-convolution layers, followed by concatenation with the corresponding copied features from the left. This is followed by 3x3 convolutions and ReLU non-linearities. 

% In the original paper, this architecture is deployed for various segmentation tasks and the experiments indicate very good performance with very few training samples, and a reasonable wall-clock training time. 

\subsection{Data-processing inequality}
% \aravind{I feel this subsection could be removed since it's just a postulation and there's no explicit analysis? The paper doesn't get weaker if this section doesn't exist imo.} \Homanga{I agree, this is mainly an hypothesis. We can put this in the Appendix if we are out of space and forward reference in the discussions.}
\label{sec:dpi}
The Data processing inequality (DPI) states that the information content of a signal cannot be increased via a local physical operation. So, given a Markov chain $X_1\rightarrow X_2 \rightarrow X_3$, the mutual information (MI) between $X_1$ and $X_2$ is not less than the MI between $X_1$ and $X_3$ i.e.
$$ MI(X_1;X_2)\geq MI(X_1;X_3)$$
A vanilla feed-forward neural network has successive layers depend only on the output of the previous layer and so there is a Markov chain of the form $X_1\rightarrow X_2 \rightarrow \cdot \cdot \cdot \rightarrow X_n$ from the input $X_1$ to the final output $X_n$. 
In practice the last layer $X_n$ contains \textit{less information} than the previous layer $X_{n-1}$. 
% Previous approaches, such as~\citep{densenet} have shown that it is difficult to train deeper feed-forward neural networks (when $n$ increases), and proposed dense (feature concatenation) connections between layers. So, we no longer have the linear Markov chain between layers when $X_k$ depends directly on a previous layer $X_{k-i}$ ($i>1$) through an identity mapping and not just on its previous layer $X_{k-1}$. Hence, the problem of DPI no longer applies when there are dense connections between layers.
% Similar residual/skip connections in the generator have also shown to stabilize training and prevent mode-collapse in VAEs~\citep{skipvaes}. The U-net architecture also incorporates information from previous layers through a copy and crop mapping between the left and right paths. 
By using dense connections~\citep{densenet}, we are able to overcome the problem of DPI, as the original input is simply concatenated with the intermediate layers of the networks.
We postulate that using dense connections are also important when parameterizing policies and value networks in RL and robotics. 
By preserving important information about the input across layers explicitly through dense connections, we can achieve faster convergence in solving complex tasks.

\subsection{Implicit under-parameterization in Deep Q-Learning}

\textcolor{black}{
\citet{aviral} showed that using MLPs for function approximation of the policy and value functions in deep RL algorithms that use bootstrapping, leads to an implicit-underparameterization phenomena that causes poor-er performance. Implicit under-parameterization refers to a reduction in the effective rank of the feature, $srank_\delta(\Phi)$ that occurs implicitly due to using MLPs for approximating Q-functions. This causes rank collapse for the feature matrix $\Phi$ which are the weights of the penultimate layer of the Q network during training.
}

\textcolor{black}{We believe that adding skip connections to the network architecture, as in D2RL could help alleviate this rank collapse issue, and improve performance. We empirically verify this in Section 5.2 and Table~\ref{tab:rank}.}

\section{Method}

In this section, we first show the issues with using deeper Multi-layered Perceptons (MLPs) to parameterize the policies and Q-networks in RL and robotics due to the Data Processing Inequality (DPI).
We then propose a simple and effective architecture which overcomes the issues, using dense-connections. 
We will denote our proposed method as: Deep Dense architectures for Reinforcement Learning or \algoName in the subsequent sections. 

% \begin{figure}[t]
%     \centering
%     \includegraphics[width=0.9\textwidth]{corl_envs-arch.png}
%     \caption{Illustrations of some of the challenging robotic control environments used for our experiments. In \textit{Fetch slide}, a Fetch robot arm with one finger must be controlled to slide a puck to a goal location. In \textit{Jacko reach}, a Jaco robot with a three finger gripper must be controlled to reach the red brick. In \textit{Ant maze}, an Ant with four legs must be controlled to navigate a maze. In \textit{Baxter block join-lift}, one arm of a Baxter robot with a two finger gripper must be controlled to join two blocks and lift the combination above a certain height.}
%     \label{fig:envs}
%     \vspace{-15pt}
% \end{figure}

\subsection{Problem with deeper networks in Reinforcement Learning}
\begin{wrapfigure}{r}{0.4\textwidth}
\vspace{-0.5cm}
  \begin{center}
    \includegraphics[width=0.4\textwidth]{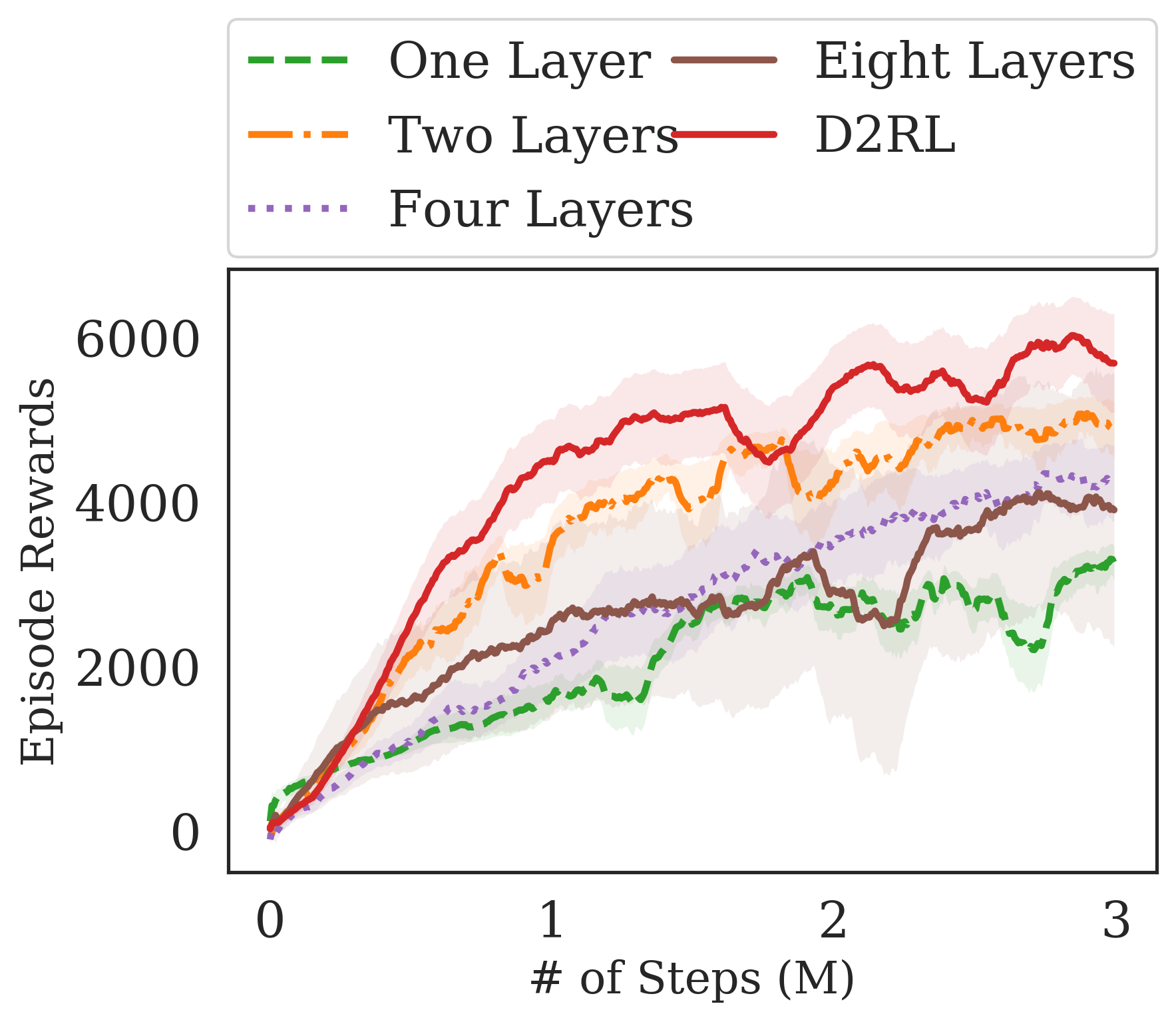}
  \end{center}
   \caption{The effect of increasing the number of fully-connected layers to parameterize the policy and Q-Networks for Soft-Actor Critic~\citep{sac} on Ant-v2 in the OpenAI Gym Suite~\citep{gym}. It is evident that performance drops when increasing depth after 2 layers. However, our \algoName agent with 4 layers does not suffer from this, and performs better.}
   %\aravind{could we have a figure that shows both the naive way of adding depth and the skipnet method? that way, it's a clear teaser figure for how we solve a problem.}\Homanga{how about now?}
%   \Animesh{right now caption reads more like ablation. Teaser caption should lead with the problem statement. 
%   We propose adding depth to policy networks. however, counter intuitively naive addition of depth layers doesnt work. We propose a novel model architecture that scales!}
%   }
    \label{fig:num_layers}
   \vspace*{-0.2cm}
\end{wrapfigure}
To test whether we can use the expressive power of deeper networks in RL for robotic control, we train a SAC agent~\citep{sac}, while increasing the number of layers in the MLP used to parameterize the policy and the Q-networks, on the Ant-v2 environment of the OpenAI Gym suite~\citep{gym}.
The results are shown in Fig. \ref{fig:num_layers}.
The results suggest that by simply increasing the number of layers used to parameterize the networks, we are unable to benefit from the inductive bias of deeper feature extractors in RL, like we see in computer vision.
As we increase the number of layers in the network, the mutual information between the output and input likely decreases due to the non-linear transformations used in deep learning as explained by the DPI (see Section \ref{sec:dpi}). 
Increasing the number of layers from 2 to 4 significantly decreases the sample efficiency of the agent and furthermore, increasing the number of layers from 2 to 8 decreases the sample efficiency while also making the agent less stable during training.
As expected, when instead decreasing the number of layers from 2 to 1 the vanilla MLPs are not sufficiently expressive to perform well on the Ant-v2 task.
This simple experiment suggests that even though the expressivity of the function approximators is important, by simply adding more layers in a vanilla MLP, the agent's performance and sample complexity significantly deteriorates. 
However, it is desirable to use deeper networks to increase the network expressivity and to enable better optimization for the networks.

% \Animesh{do we need the teaser figure in ICLR paper. can we move fig 1 here.?? }
\subsection{D2RL}

Our proposed \algoName variant incorporates dense connections (input concatenations) from the input to each of the layers of the MLP used to parameterize the policy and value functions. We simply concatenate the state or the state-action pair to each hidden layer of the networks except the last output linear layer, since that is just a linear transformation of the output from the previous layer. In case of pixel observations, we consider the states to be the encodings of a CNN encoder, as shown in Fig.~\ref{fig:teaser}. Our simple architecture enables us to increase the depth of the networks while also satisfying DPI.
Fig \ref{fig:teaser} is a visual illustration and we provide a PyTorch-like pseudo-code below to promote clarity of the proposed method.
We also include the actual PyTorch code~\citep{pytorch} for the policy and value networks to allow for fast-adoption and reproducibility in Appendix \ref{pytorch_code}.

% \aravind{could we just use MLP() instead of activation(linear())? that way, we can avoid writing linear1, linear2, etc.}\Homanga{fixed.}
\begin{lstlisting}[language=Python,label={lst:pytorch_alg}]
# Sample state, action from the replay buffer
state, action = replay_buffer.sample()
# Feed state, action into the first linear layer of a Q-network
q_input = concatenate(state, action)
h = MLP(q_input)
# Concatenate the hidden representation with the input
h = concatenate(h, q_input)
# Feed the concatenated representation into the second layer
h = MLP(h)
\end{lstlisting}
\vspace{-3pt}

% We consider actor-critic RL algorithms and propose to consider policy and value functions parameterized as shown in Fig.~\ref{fig:mainfig}. This architecture is very similar to the original U-Net architecture and has all the three primary components - a contractive path on the left, an expansive path on the right, and copy+crop (skip) connections from the left path to the right path.

% We use a similar architecture for both the policy and Q-value networks. We perform ablations by making this architecture change in only the policy network and only the value network, the results for which are in the Appendix. 

\section{Experiments}

We experiment across a suite of different environments, some of which are shown in Fig.~\ref{fig:envs}, each simulated using the MuJoCo simulator. We were unable to do real robot experiments due to COVID-19 and have included a 1-page statement along with the Appendix describing how our method can conveniently scale to physical robots. 
Through the experiments, we aim to understand the following questions:
\begin{itemize}
    \item How does \algoName compare with the baseline algorithms in terms of both asymptotic performance and sample efficiency on challenging robotic environments?
    % \item Is the \algoName variation more sample efficient than the baseline algorithms?
    \item Are the benefits of \algoName consistent across a diverse set of algorithms and environments?
    \item Is \algoName important for both the policy and value networks? How does \algoName perform with increasing depth? %(Ablations) \aravind{also mention ablation study analysis} \Homanga{fixed.}
\end{itemize}

\subsection{Experimental details}
%\Homanga{TODO} In this section we discuss the exact experimental settings and the baselines. All architecture details (except hyperparams) go here.

% \aravind{it would be nice to just make a single Table, where we mention state-based or pixel-based, what is the underlying RL algorithm, what is the architecture for the baseline and for our model, number of parameters for each of them, and performance. that way, people can easily identify what is going on. Once we do that, it would become much more easy to say there's no other change anywhere else than the architecture. Headers can be as follows: Method (baseline / R3L); Type of input (state / pixel); Architecture (4-layer MLP or SkipNet) ; RL algorithm  (SAC / TD3 / CURL); Score (100K); Score (500k), Score (1M).}

% \aravind{Similarly, it would be nice to just make a single Table for all the environments, their properties, ex Env, Input Type (state / pixel), Task type (loco/manip), RL algo for baseline (SAC/HER/HIRO/TD3/CURL)}

% \aravind{the above two tables can save you space by removing the verbosity in the experiment section, and you could use the space to focus more on the insights and conclusions from the results, what question is asked, what is the answer from experiment, what is the take-away. also present the questions ahead in the beginning of the section and reference to them.} \Homanga{I feel we should do this arrangement in the form of tables, but in the Appendix, as we still want to have plots in the main paper and minimal details about the environment/hyperparams.}

To enable fair comparison between the current standard baselines and \algoName, we simply replace the 2-layer MLPs that are commonly parameterize the policy and value function(s) in widely used actor-critic algorithms such as SAC~\citep{sac, sac_ae}, TD3~\citep{td3}, DDPG~\citep{ddpg} and HIRO~\citep{hiro}. Instead, we use 4-layer \algoName to parameterize both the policies and the value function(s) in each of the actor-critic algorithms.
Outside of the network architecture, we \textbf{do not change any hyperparameters}, and use the exact same values as reported in the original papers and the corresponding open-source code repositories. The details of all the hyperparameters used are in Appendix \ref{hyperparams}. We perform ablation studies in Section \ref{sec:ablation} to $i)$ investigate the importance of parameterizing both the policy and value function(s) using \algoName and $ii)$ see how varying the number of layers of \algoName affects performance.
We also perform further experiments with a ResNet style architecture and additional experiments with Hindsight Experience Replay~\citep{her} on simpler manipulation environments which can be found in Appendix \ref{additional_exps}.

\subsection{Experimental Results}
\begin{figure*}[t]
\centering
    \begin{subfigure}[b]{0.19\textwidth}
        \centering
        \includegraphics[width=\textwidth]{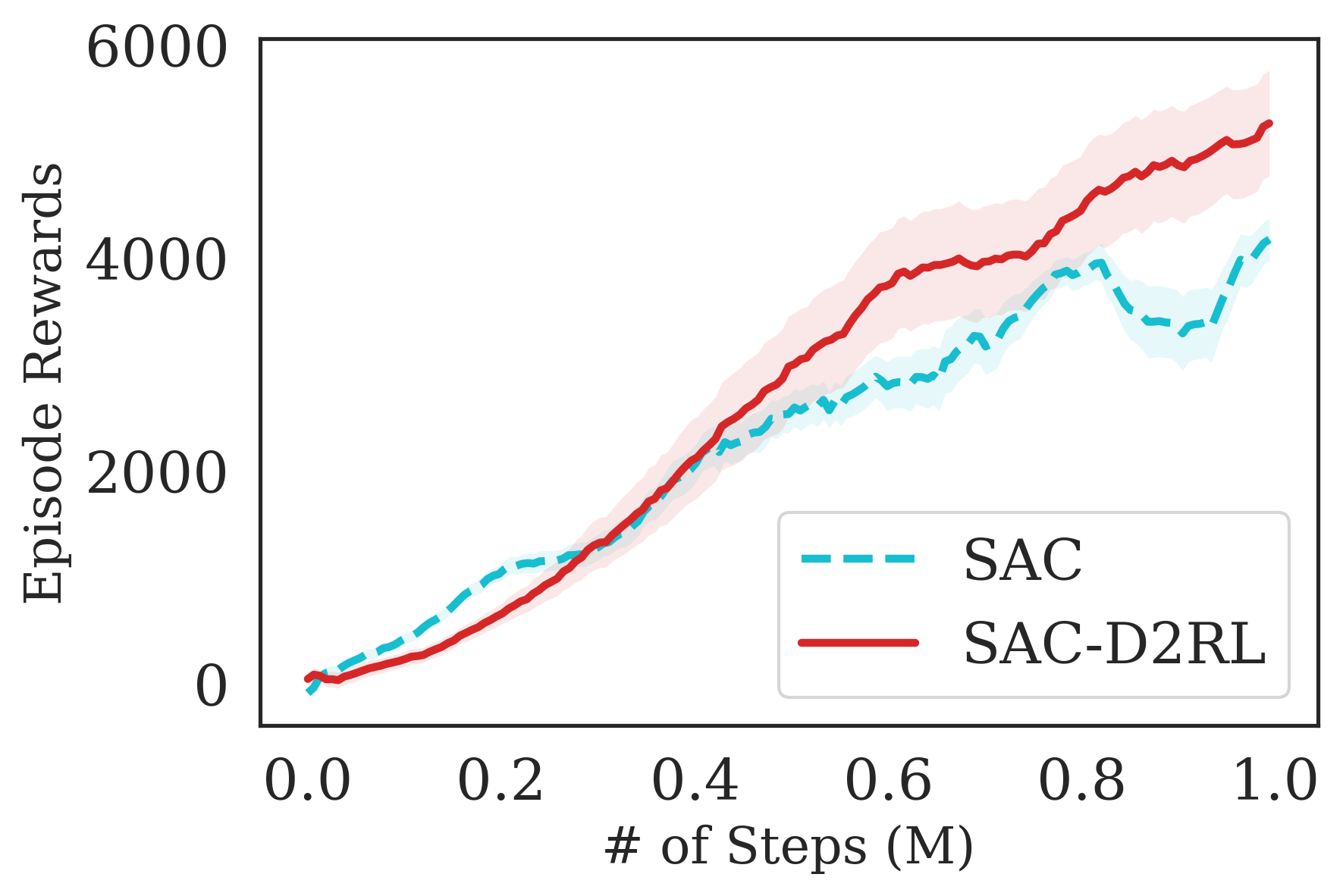}
        \caption{Gym Ant SAC}
        \label{fig:ant_sac}
    \end{subfigure}% \\
    \begin{subfigure}[b]{0.19\textwidth}
        \centering
        \includegraphics[width=\textwidth]{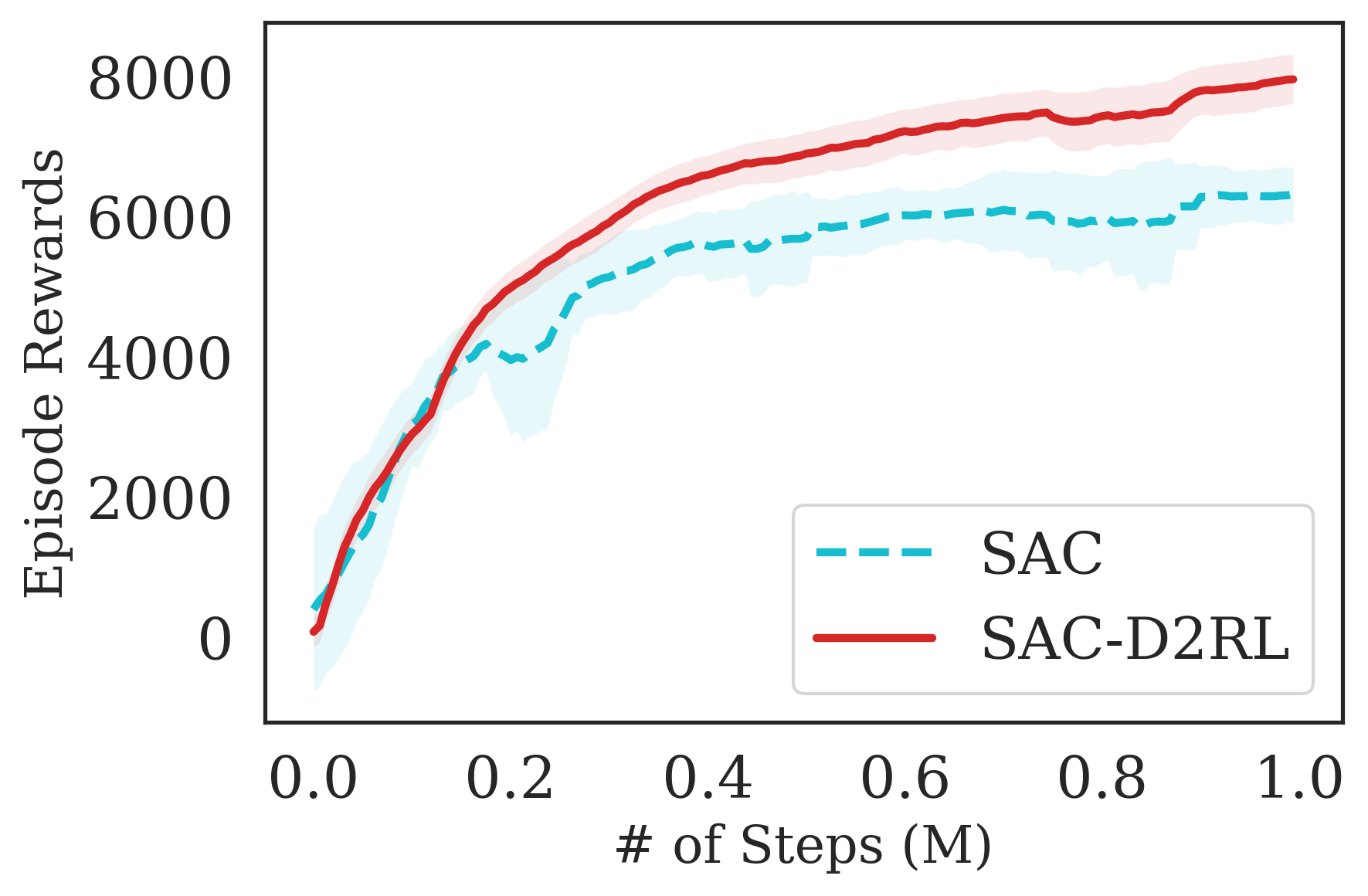}
        \caption{Gym Cheetah SAC}
        \label{fig:cheetah_sac}
        \end{subfigure}
    \begin{subfigure}[b]{0.19\textwidth}
        \centering
        \includegraphics[width=\textwidth]{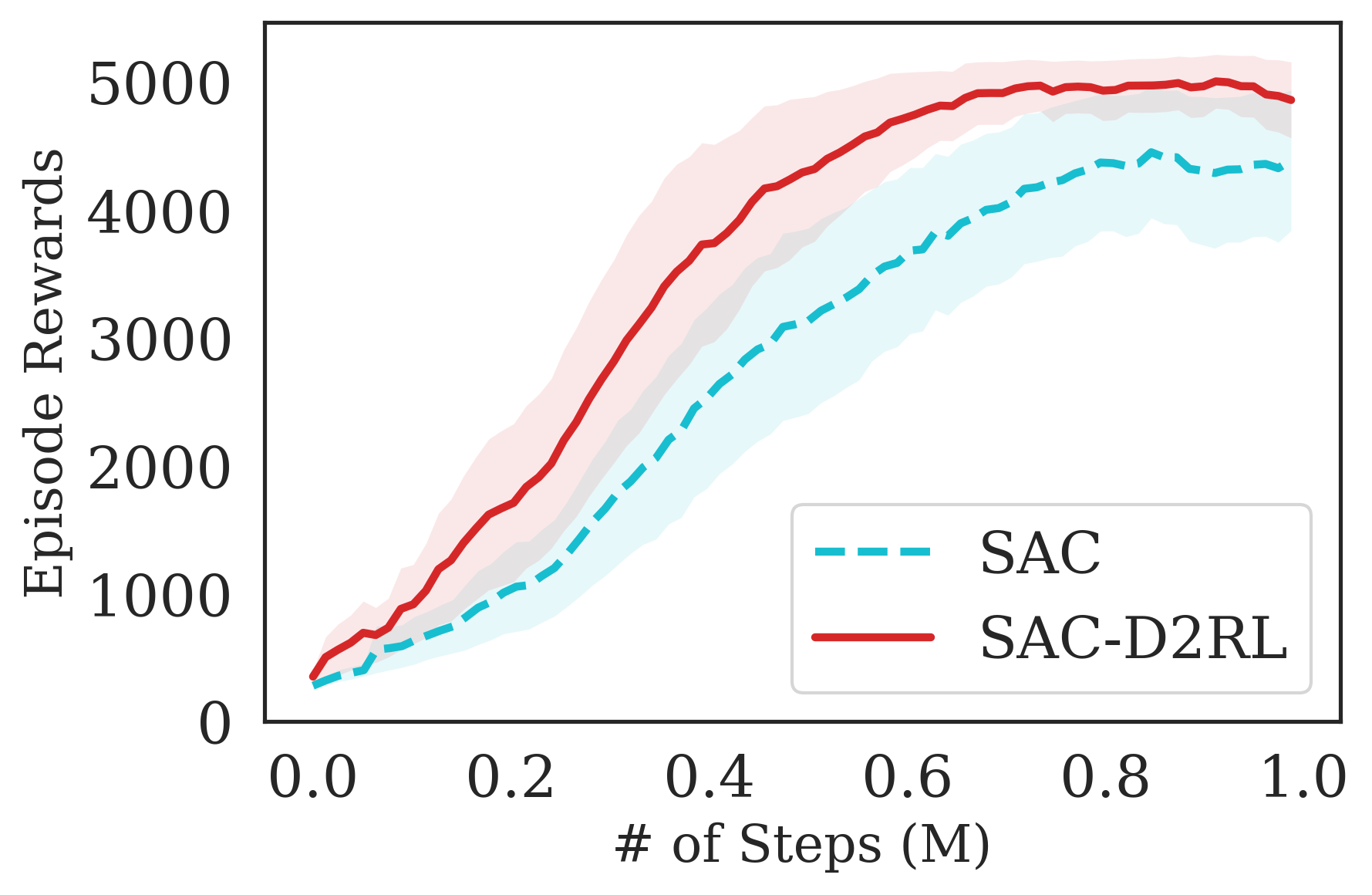}
        \caption{Gym Humanoid SAC}
        \label{fig:hopper_sac}
    \end{subfigure}
    \begin{subfigure}[b]{0.19\textwidth}
        \centering
        \includegraphics[width=\textwidth]{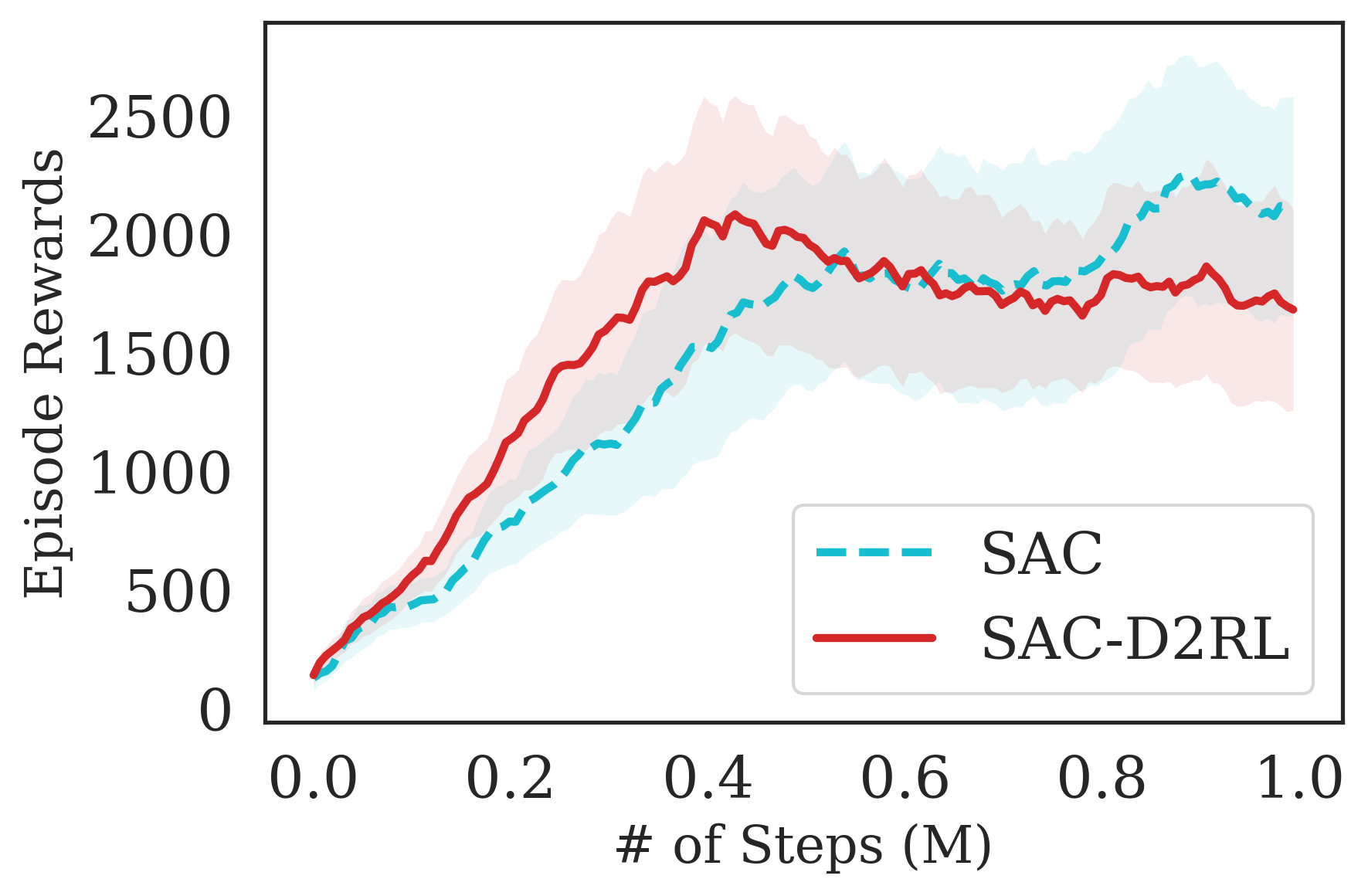}
        \caption{Gym Hopper SAC}
        \label{fig:hopper_sac}
    \end{subfigure}
    % \begin{subfigure}[b]{0.33\textwidth}
    %     \centering
    %     \includegraphics[width=\textwidth]{plots/humanoid_sac.png}
    %     \caption{Gym Humanoid SAC}
    %     \label{fig:humanoid_sac}
    % \end{subfigure}
    \begin{subfigure}[b]{0.19\textwidth}
        \centering
        \includegraphics[width=\textwidth]{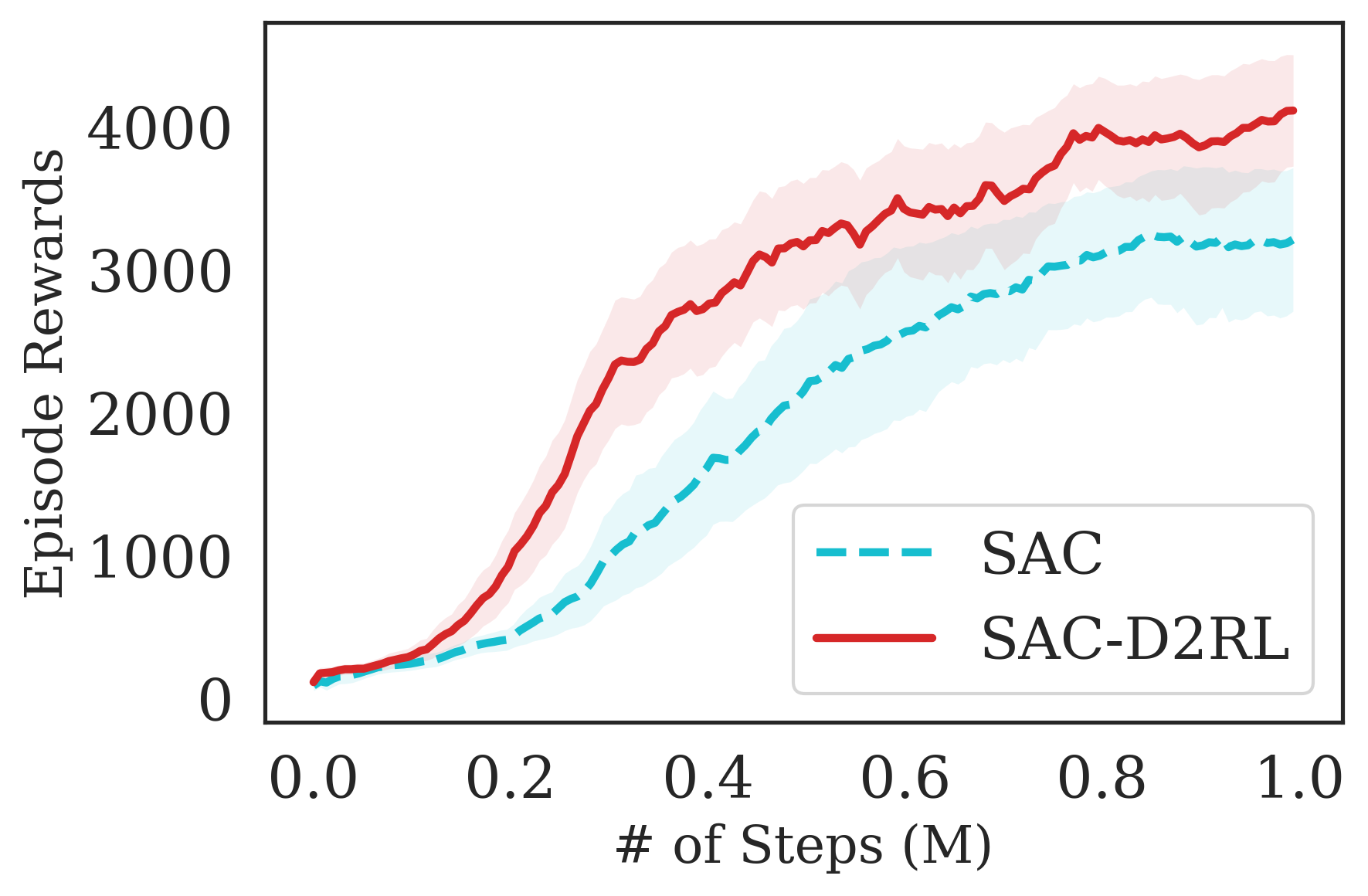}
        \caption{Gym Walker SAC}
        \label{fig:walker_sac}
    \end{subfigure}
    %      \begin{subfigure}[b]{0.2\textwidth}
    %           \centering
    % \includegraphics[width=\textwidth]{pushnreach.png}
    % \caption{Sawyer Push and Reach (Image)}
    % \label{fig:pushreach}
    %     \end{subfigure}
    %\quad
     \caption{\textbf{OpenAI Gym benchmark environments with SAC.} Comparison of the proposed \algoName and the baselines on a suite of OpenAI-Gym environments. We apply the \algoName modification to SAC~\citep{sac}. The error bars are with respect to 5 random seeds. The results on Humanoid env are in the Appendix.}
    %  \Animesh{replace hopper with humanoid in main paper -- better results -- I guess its not here since we dont have symmetric humanoid in TD3?}}  
 \label{fig:sim_resultsgym}
\end{figure*}
\begin{figure*}[t]
\centering
    \begin{subfigure}[b]{0.24\textwidth}
        \centering
        \includegraphics[width=\textwidth]{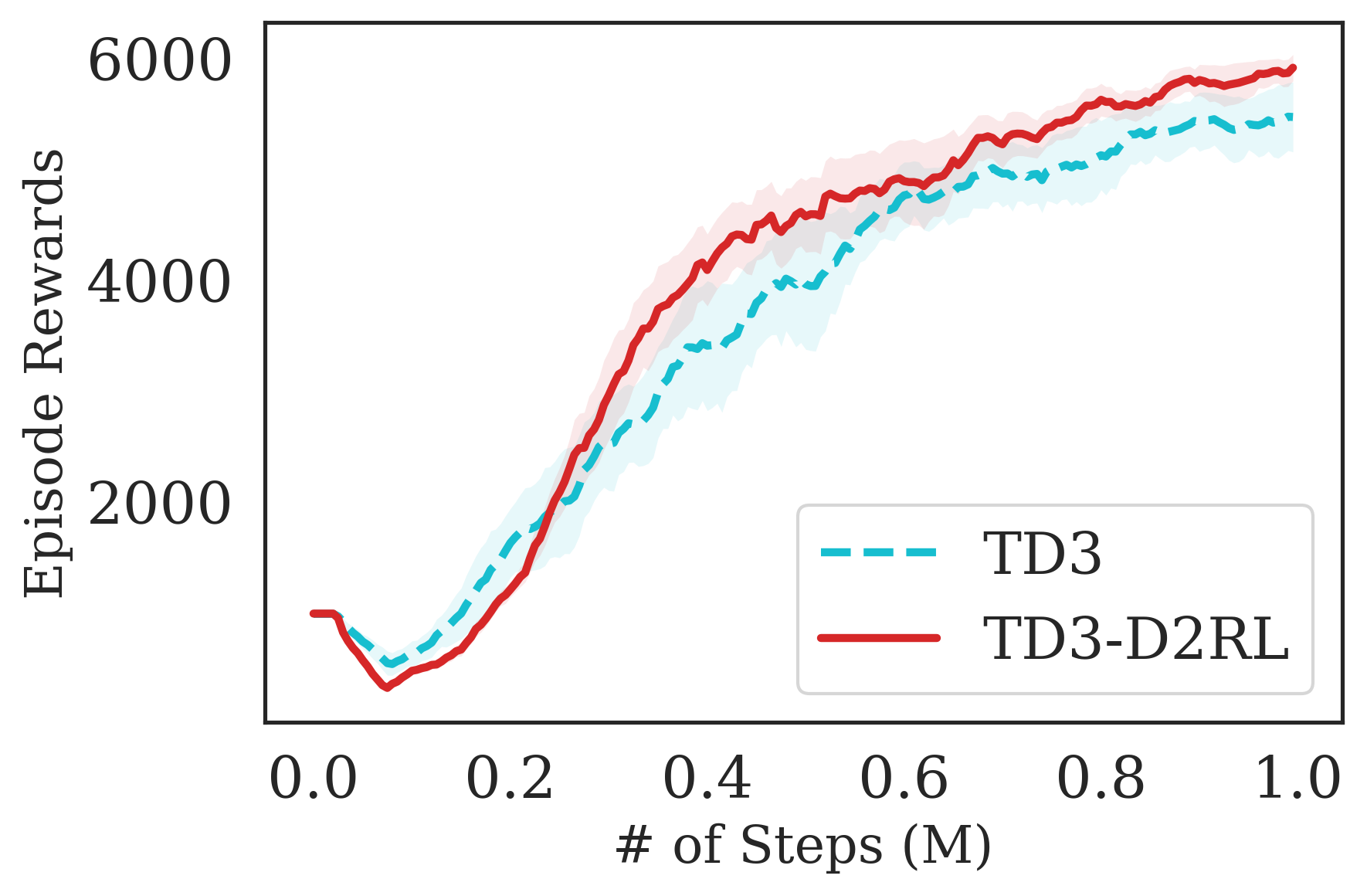}
        \caption{Gym Ant TD3}
        \label{fig:ant_td3}
    \end{subfigure}
    \begin{subfigure}[b]{0.24\textwidth}
        \centering
        \includegraphics[width=\textwidth]{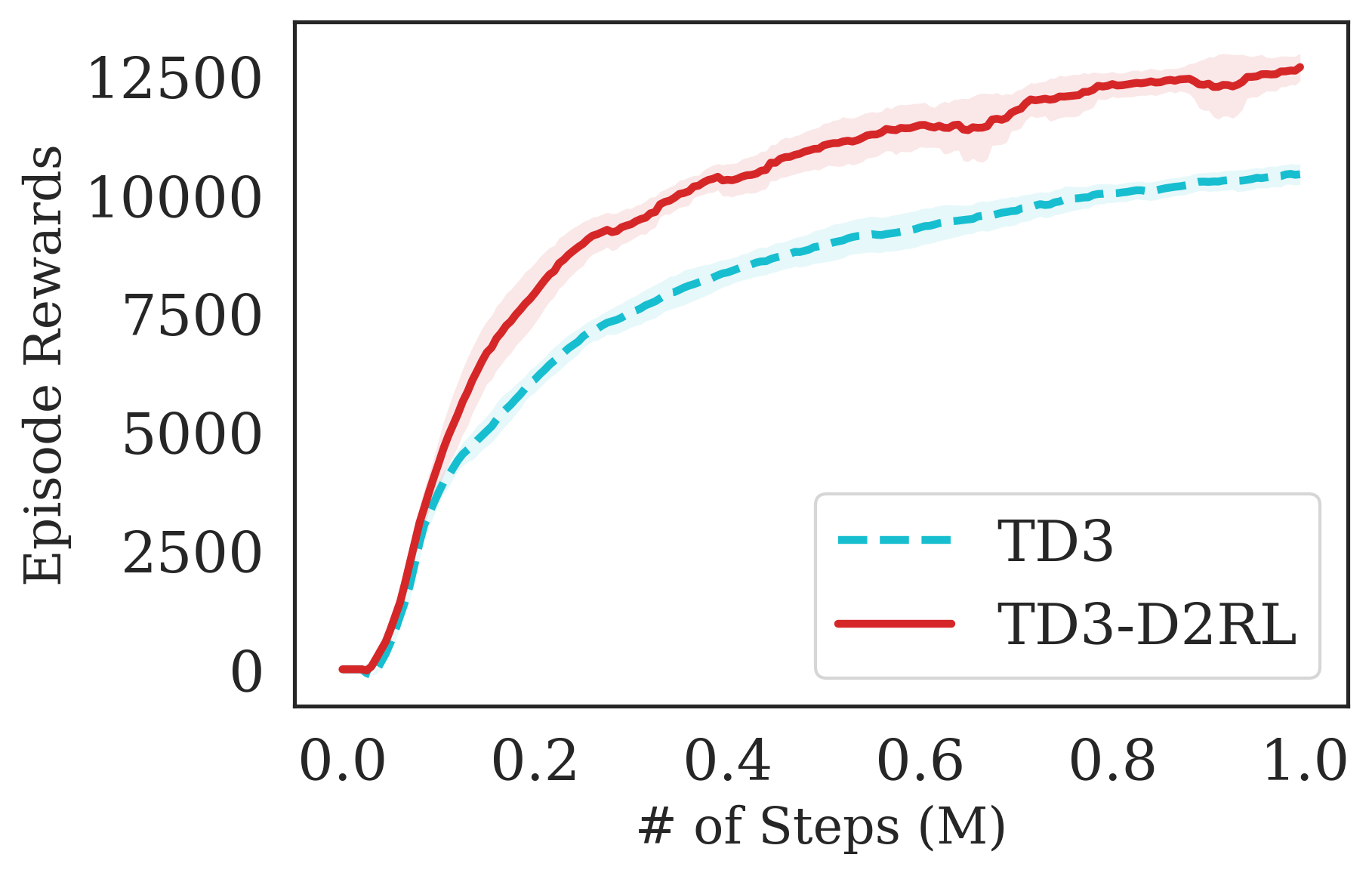}
        \caption{Gym Cheetah TD3}
        \label{fig:cheetah_td3}
    \end{subfigure}
    \begin{subfigure}[b]{0.24\textwidth}
        \centering
        \includegraphics[width=\textwidth]{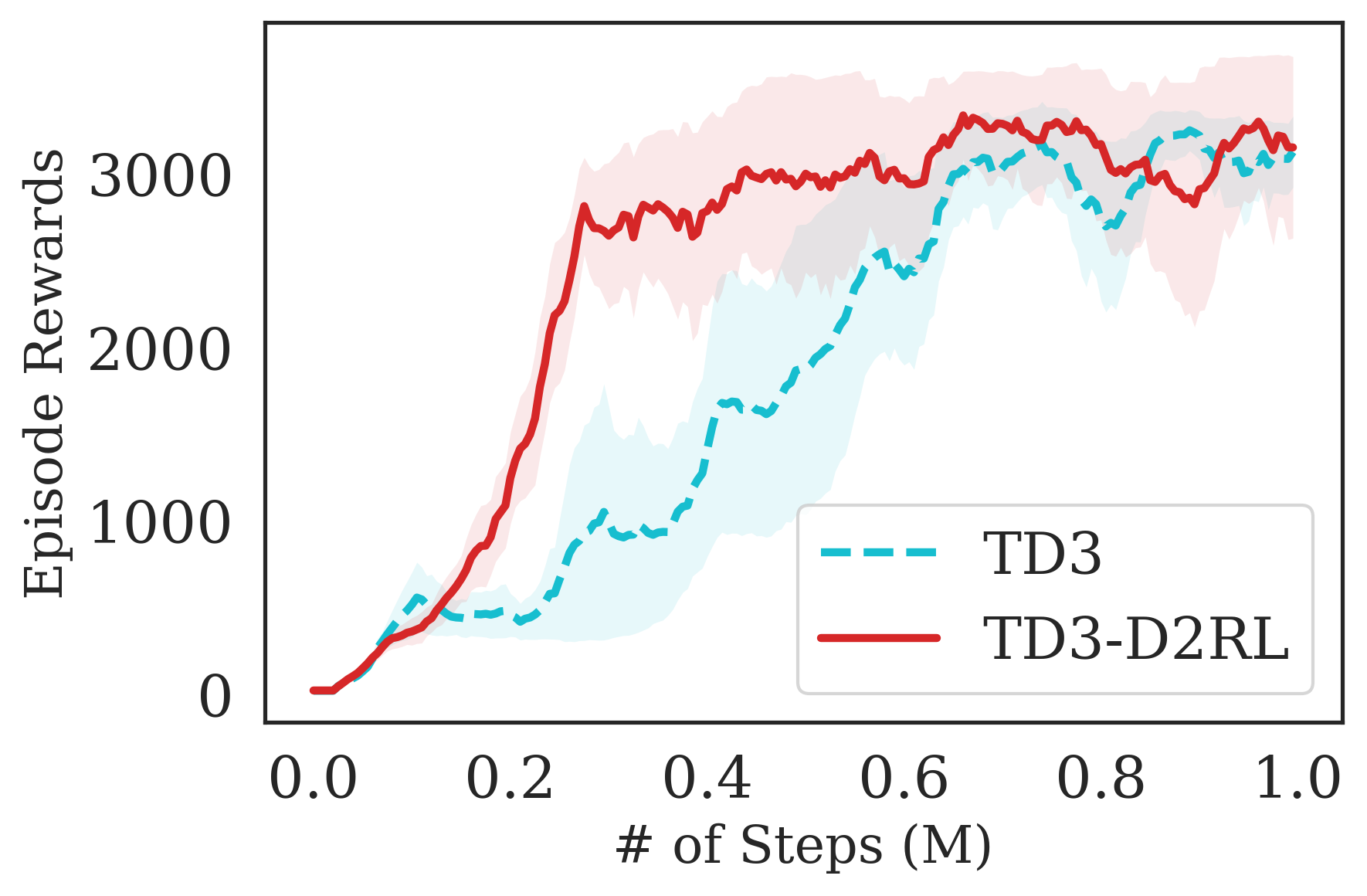}
        \caption{Gym Hopper TD3}
        \label{fig:hopper_td3}
    \end{subfigure}
    \begin{subfigure}[b]{0.24\textwidth}
        \centering
        \includegraphics[width=\textwidth]{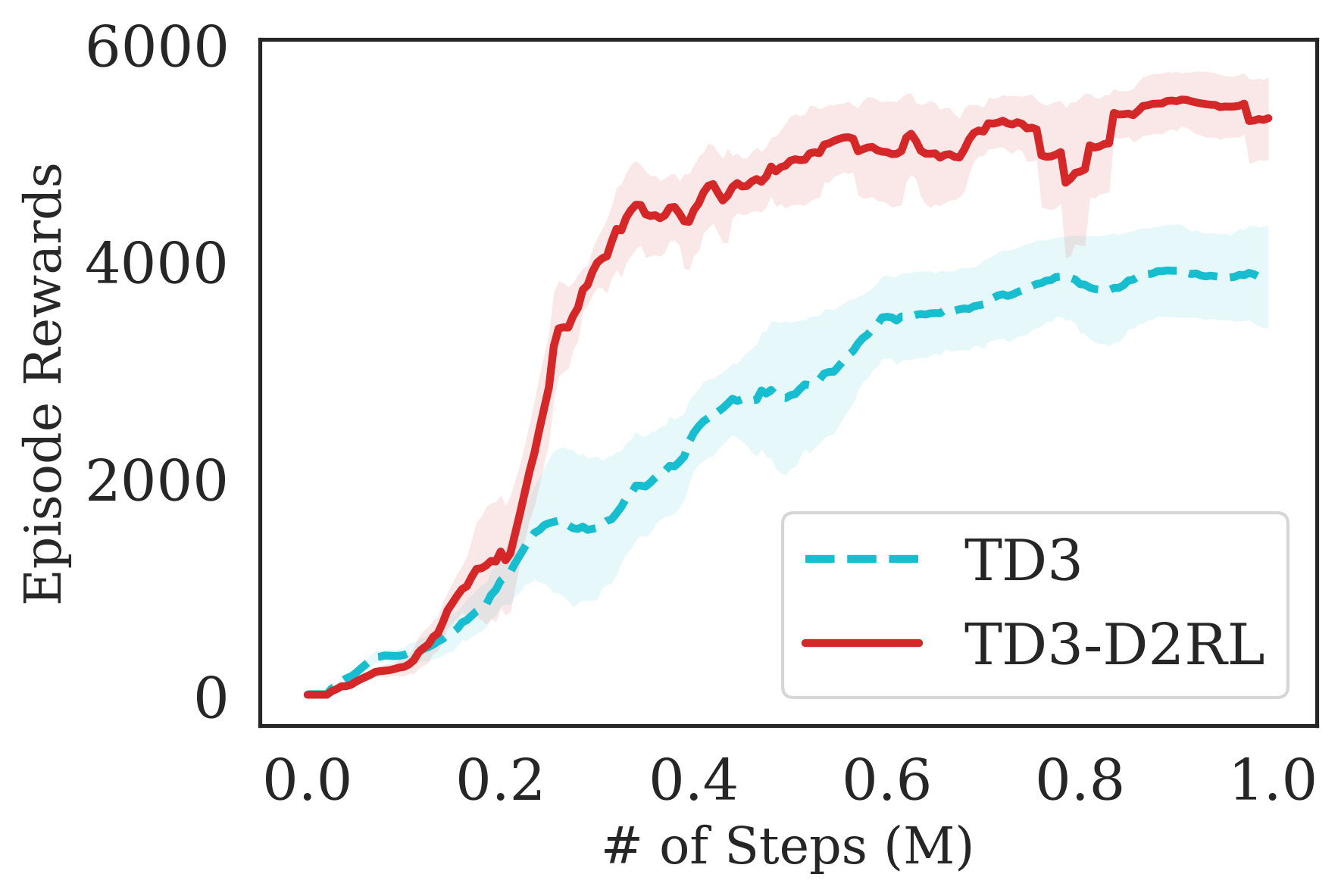}
        \caption{Gym Walker TD3}
        \label{fig:walker_td3}
    \end{subfigure}
\caption{\textbf{OpenAI Gym benchmark environments with TD3.} Comparison of the proposed variation \algoName and the baselines on a suite of OpenAI-Gym environments. We apply the \algoName modification to TD3~\citep{td3}. The error bars are with respect to 5 random seeds.}  
 \label{fig:sim_resultsgym}
\end{figure*}

\noindent\textbf{The proposed \algoName variant achieves superior performance compared to the baselines on state-based OpenAI Gym MuJoCo environments.} We benchmark the proposed \algoName variant on a suite of OpenAI Gym~\citep{gym} environments by applying the modification to two standard RL algorithms, SAC~\citep{sac} and TD3~\citep{td3}. For all the environments, namely Ant, Cheetah, Hopper, Humanoid, and Walker, the observations received by the agent are state vectors consisting of the positions and velocities of joints of the agent. Additional details about the state-space and action-space are in the Appendix.
From Fig.~\ref{fig:sim_resultsgym}, we see that the proposed \algoName modification converges to significantly higher episodic rewards in most environments, and in others is competitive with the baseline. Also, these benefits can be seen across both the algorithms, SAC and TD3.

 \begin{table*}[t]
 \color{black}
 \small
   \setlength\tabcolsep{1.6pt}
   \centering
%   \begin{minipage}[c]{0.65\textwidth}
  \resizebox{0.95\linewidth}{!}{
  \centering
   \begin{tabular}{l | ccc|ccc}
     \toprule
      & \multicolumn{3}{c|}{\textbf{From Images}} & \multicolumn{3}{c}{\textbf{From States}}  \\
     \midrule
    \textbf{100K Step score}  & \textbf{CURL} &
    \textbf{CURL-ResNet} &\textbf{CURL-\algoName} & \textbf{SAC} &
    \textbf{SAC-ResNet}& \textbf{SAC-\algoName} \\
    \midrule
    Finger, Spin & 767 $\pm$ 57 & 548 $\pm$ 120&\textbf{837} $\pm$ 18 & 459 $\pm$ 48 & 551 $\pm$ 57 &\textbf{627} $\pm$ 107 \\
    \rowcolor[HTML]{EFEFEF}
    Cartpole, Swing & 582 $\pm$ 142 & 327 $\pm$ 101 & \textbf{836} $\pm$ 34 & 717 $\pm$ 14 & 701 $\pm$ 24 &\textbf{751} $\pm$ 12 \\
    Reacher, Easy & 538 $\pm$ 233 & 526 $\pm$ 79 &\textbf{754} $\pm$ 168 & \textbf{752} $\pm$ 121 & 626 $\pm$ 112 & 675 $\pm$ 203 \\
    \rowcolor[HTML]{EFEFEF}
    Cheetah, Run & \textbf{299} $\pm$ 48  & 230 $\pm$ 17 & 253 $\pm$ 57 & 587 $\pm$ 58 & 633 $\pm$ 28 & \textbf{721} $\pm$ 43 \\
    Walker, Walk & 403 $\pm$ 24 & 275 $\pm$ 57& \textbf{540} $\pm$ 143 & 132 $\pm$ 43 & \textbf{456} $\pm$ 89 & \textbf{354} $\pm$ 159 \\
    \rowcolor[HTML]{EFEFEF}
    Ball in Cup, Catch & 769 $\pm$ 43 & 450 $\pm$ 169 & \textbf{880} $\pm$ 48 & 867 $\pm$ 42 & \textbf{880} $\pm$ 22 & \textbf{891} $\pm$ 33 \\
    \midrule
    \textbf{500K Step score}   & \textbf{CURL} & \textbf{CURL-ResNet} & \textbf{CURL-\algoName} & \textbf{SAC} & \textbf{SAC-ResNet} & \textbf{SAC-\algoName} \\
    \midrule
    Finger, Spin & 926 $\pm$ 45 & 896 $\pm$ 59 & \textbf{970} $\pm$ 14 & 899 $\pm$ 29 & 917 $\pm$ 21 & \textbf{961} $\pm$ 8 \\
    \rowcolor[HTML]{EFEFEF}
    Cartpole, Swing & 841 $\pm$ 45 & 833 $\pm$ 9 & \textbf{859} $\pm$ 8 & 884 $\pm$ 2 & 885 $\pm$ 4 &  885 $\pm$ 2 \\
    Reacher, Easy & \textbf{929} $\pm$ 44 & 900 $\pm$ 48& \textbf{929} $\pm$ 62 & \textbf{973} $\pm$ 23 & \textbf{969} $\pm$ 34 & 952 $\pm$ 30  \\
    \rowcolor[HTML]{EFEFEF}
    Cheetah, Run & \textbf{518} $\pm$ 28 & 459 $\pm$ 108 & 386 $\pm$ 115 & 781 $\pm$ 65 & 807 $\pm$ 69 & \textbf{842} $\pm$ 75\\
    Walker, Walk & 902 $\pm$ 43 & 807 $\pm$ 134 & \textbf{931} $\pm$ 24 & \textbf{959} $\pm$ 16 & \textbf{972} $\pm$ 8 &  \textbf{964} $\pm$ 14 \\
    \rowcolor[HTML]{EFEFEF}
    Ball in Cup, Catch & 959$\pm$ 27 &  960$\pm$ 8 &955 $\pm$ 15 & 976 $\pm$ 12 & 970 $\pm$ 19 & 972 $\pm$ 13 \\
    \bottomrule
    \end{tabular}
    }
    %   \end{minipage}
    %   \hspace*{-0.3cm}
    % \begin{minipage}[c]{0.33\textwidth}
    \caption{\textbf{DeepMind control suite \textit{benchmark} environments from images (CURL) and states (SAC).} Results of CURL~\citep{curl}, CURL-\algoName, SAC~\citep{sac}, and SAC-\algoName, on the standard DM Control Suite \textit{benchmark} environments. CURL~\citep{curl} and CURL-\algoName are trained purely with pixel observations while SAC~\citep{sac} and SAC-\algoName are trained with proprioceptive features. The results for CURL were taken directly as reported by \citet{curl}. \textcolor{black}{CURL-ResNet is the baseline that uses a similar network as CURL-\algoName but with residual connections. Similarly, SAC-ResNet is the baseline that uses a similar network as SAC-\algoName but with residual connections. } The S.D. is over 5 random seeds.}
    \label{tab:DML_resutls}
    % \end{minipage}
   % \vspace{-15pt}
 \end{table*}

\begin{table*}[t!]
 \small
   \setlength\tabcolsep{1.6pt}
   \centering
%   \resizebox{0.95\linewidth}{!}{
  \centering
\begin{tabular}{l c c|l c c}
    \toprule
    \textbf{100K Step Score} & \textbf{TD3} & \textbf{TD3-\algoName} & \textbf{500K Step Score}& \textbf{TD3} & \textbf{TD3-\algoName} \\
    \midrule
    Walker, Walk-Easy & \textbf{67} $\pm$ 8 & \textbf{69} $\pm$ 11 & Walker, Walk-Easy & \textbf{79} $\pm$ 10 & \textbf{77} $\pm$ 8 \\
    \rowcolor[HTML]{EFEFEF}
    Walker, Walk-Medium & 38 $\pm$ 8 & \textbf{60} $\pm$ 9 & Walker, Walk-Medium & 44 $\pm$ 8  & \textbf{69} $\pm$ 6 \\
    Cartpole, Swing-Easy & 96 $\pm$ 24 & \textbf{142} $\pm$ 18 & Cartpole, Swing-Easy & 102 $\pm$ 24 & \textbf{171} $\pm$ 32  \\
    \rowcolor[HTML]{EFEFEF}
    Cartpole, Swing-Medium & 67 $\pm$ 2 & \textbf{102} $\pm$ 13 & Cartpole, Swing-Medium & 68 $\pm$ 2 & \textbf{138} $\pm$ 16 \\
    \bottomrule
 \end{tabular}
    % }
       \hspace*{-0.3cm}
    % \begin{minipage}[c]{0.25\textwidth}
    \caption{\textbf{Real World RL suite environments from states.} Results of TD3~\citep{sac}, and TD3-\algoName, on the Real World RL suite environments after 500K environment steps over 5 seeds.  We see that using \algoName we are able to 
    perform better in environments with distractors, random noise and delays.
    These experiments show how \algoName is able to learn robust agents.}
    \label{tab:RWRLresults}
    % \end{minipage}
    \vspace{-0.1in}
\end{table*}
 
\begin{figure*}[t!]
\centering
    % \begin{subfigure}[b]{0.3\textwidth}
    %     \centering
    %     \includegraphics[width=\textwidth]{plots/pick_her.png}
    %     \caption{Fetch Pick HER}
    %     \label{fig:pick_her}
    % \end{subfigure}% \\
    % \begin{subfigure}[b]{0.3\textwidth}
    %     \centering
    %     \includegraphics[width=\textwidth]{plots/push_her.png}
    %     \caption{Fetch Push HER}
    %     \label{fig:push_her}
    % \end{subfigure}
      \begin{subfigure}[b]{0.32\textwidth}
        \centering
        \includegraphics[width=\textwidth]{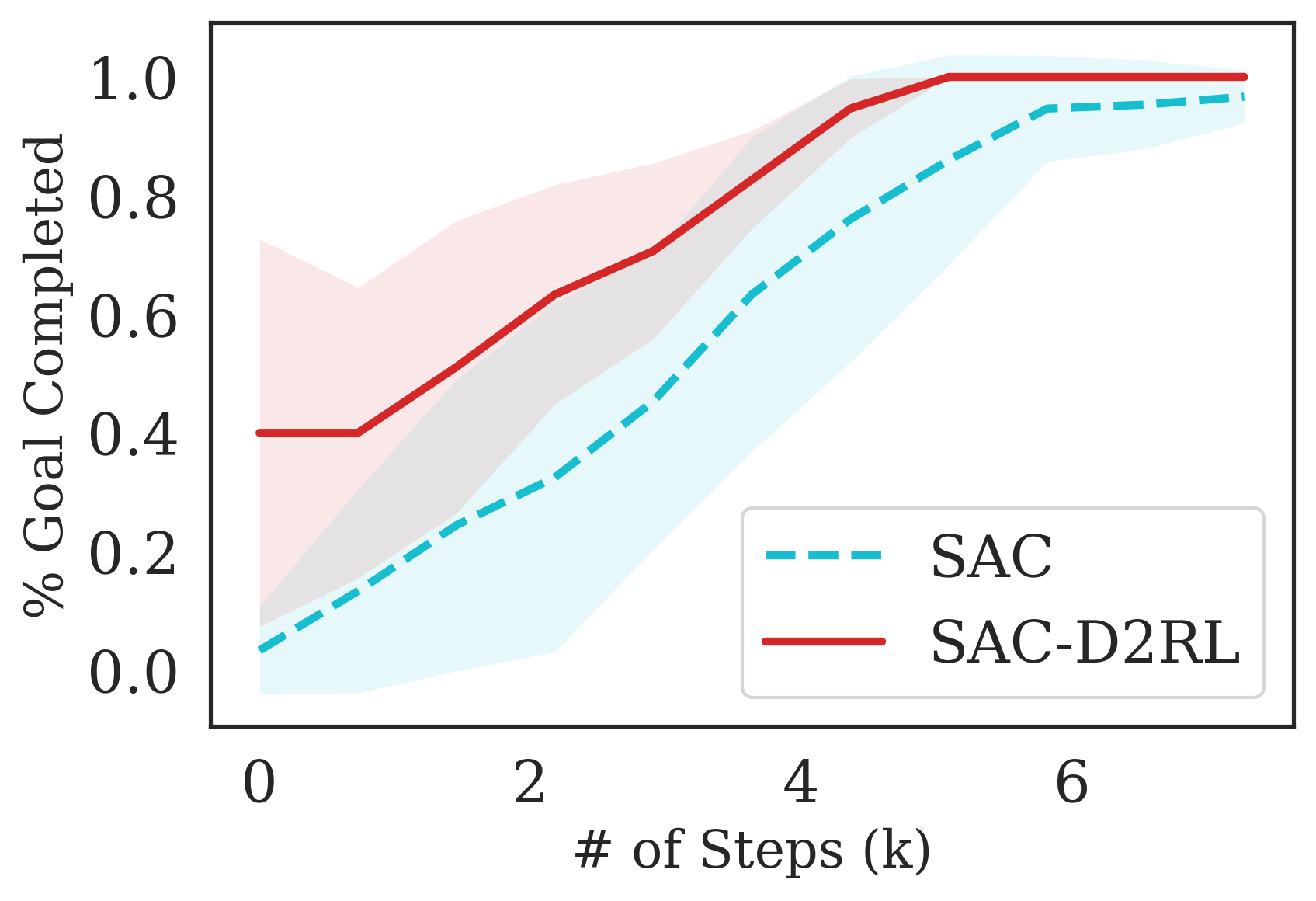}
        \caption{Fetch Reach SAC}
        \label{fig:reach_sac}
    \end{subfigure}
        \begin{subfigure}[b]{0.32\textwidth}
        \centering
        \includegraphics[width=\textwidth]{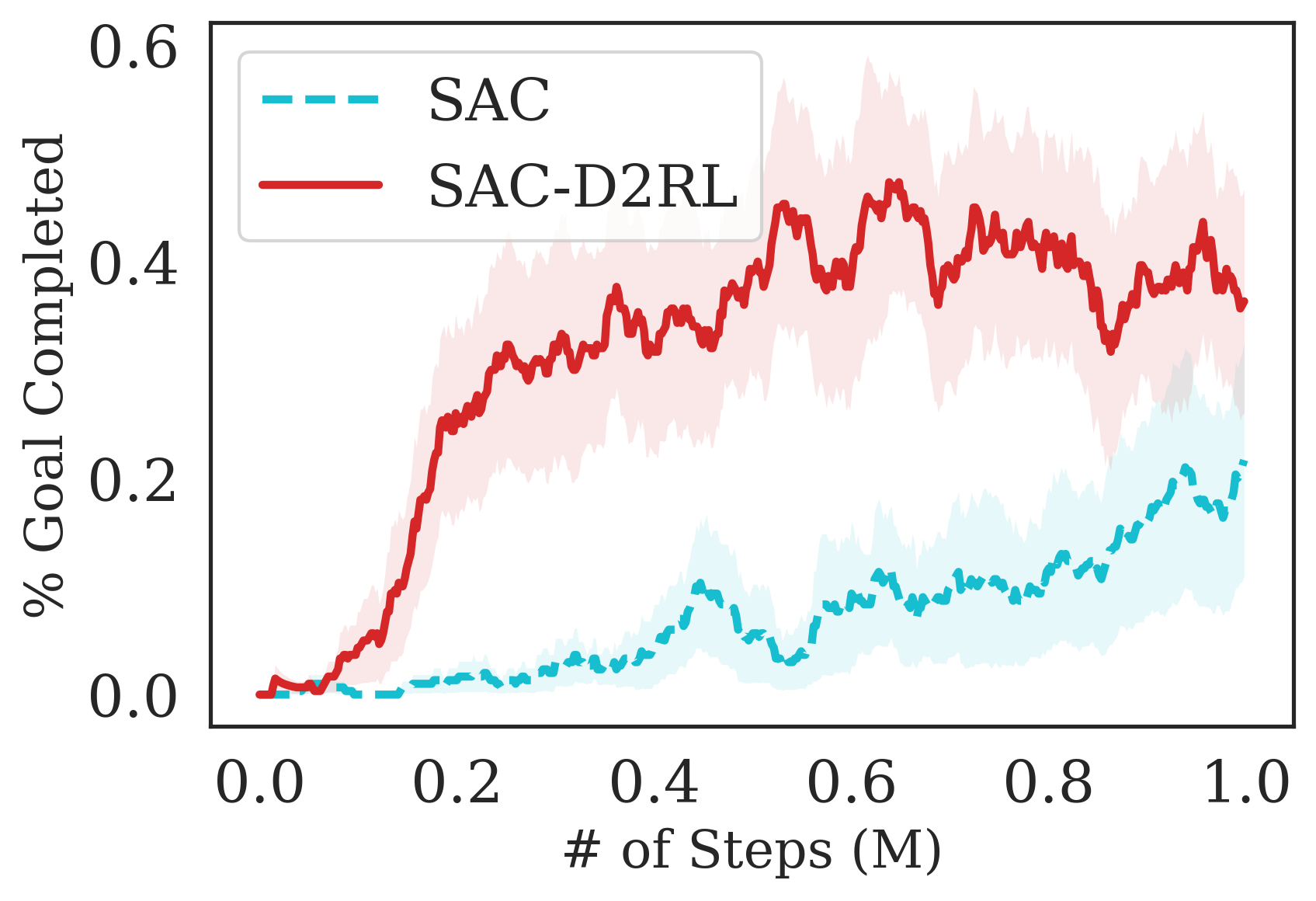}
        \caption{Fetch Slide SAC}
        \label{fig:slide_sac}
    \end{subfigure}
      \begin{subfigure}[b]{0.32\textwidth}
        \centering
        \includegraphics[width=\textwidth]{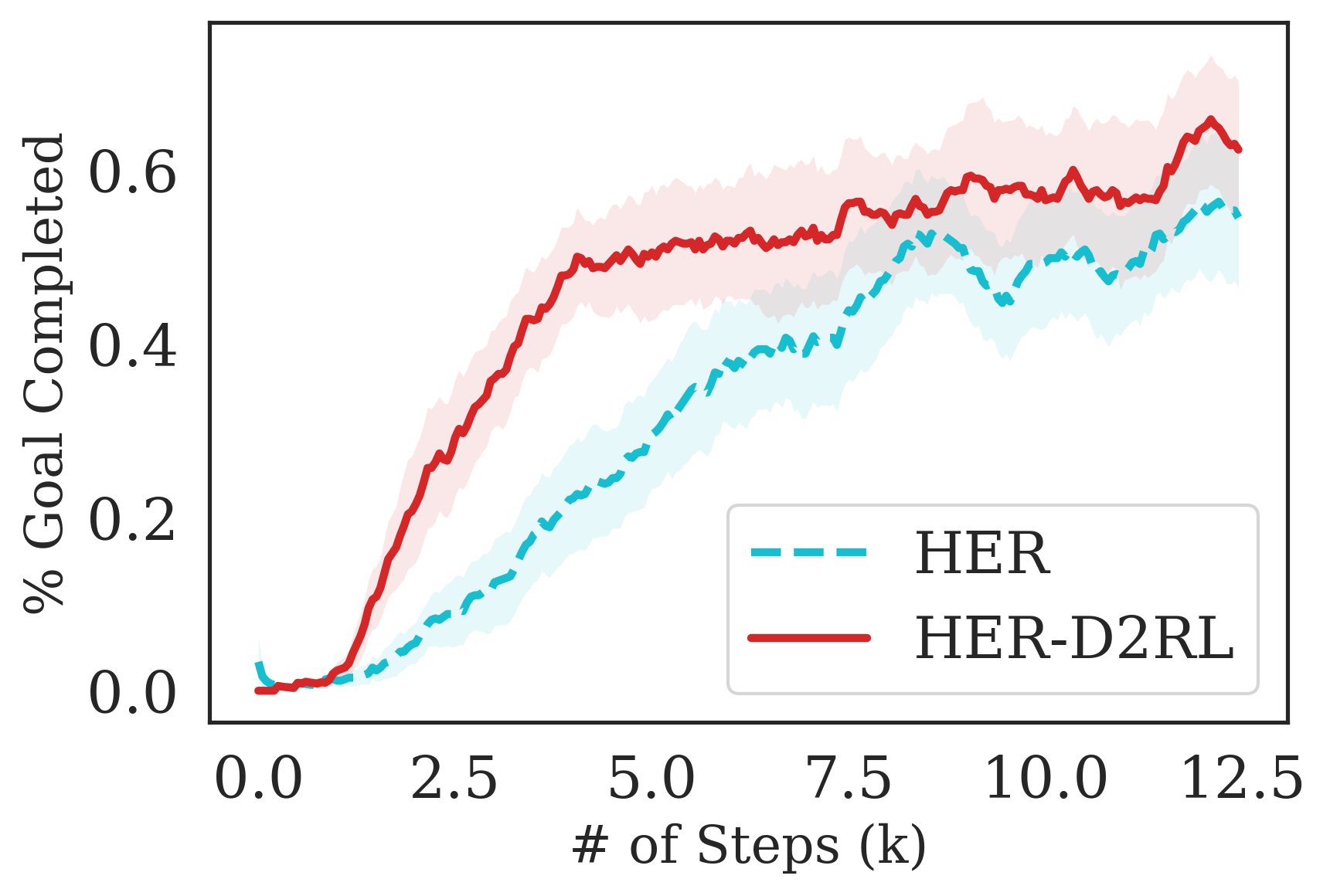}
        \caption{Fetch Slide HER}
        \label{fig:narrow}
    \end{subfigure}
    \begin{subfigure}[b]{0.32\textwidth}
        \centering
        \includegraphics[width=\textwidth]{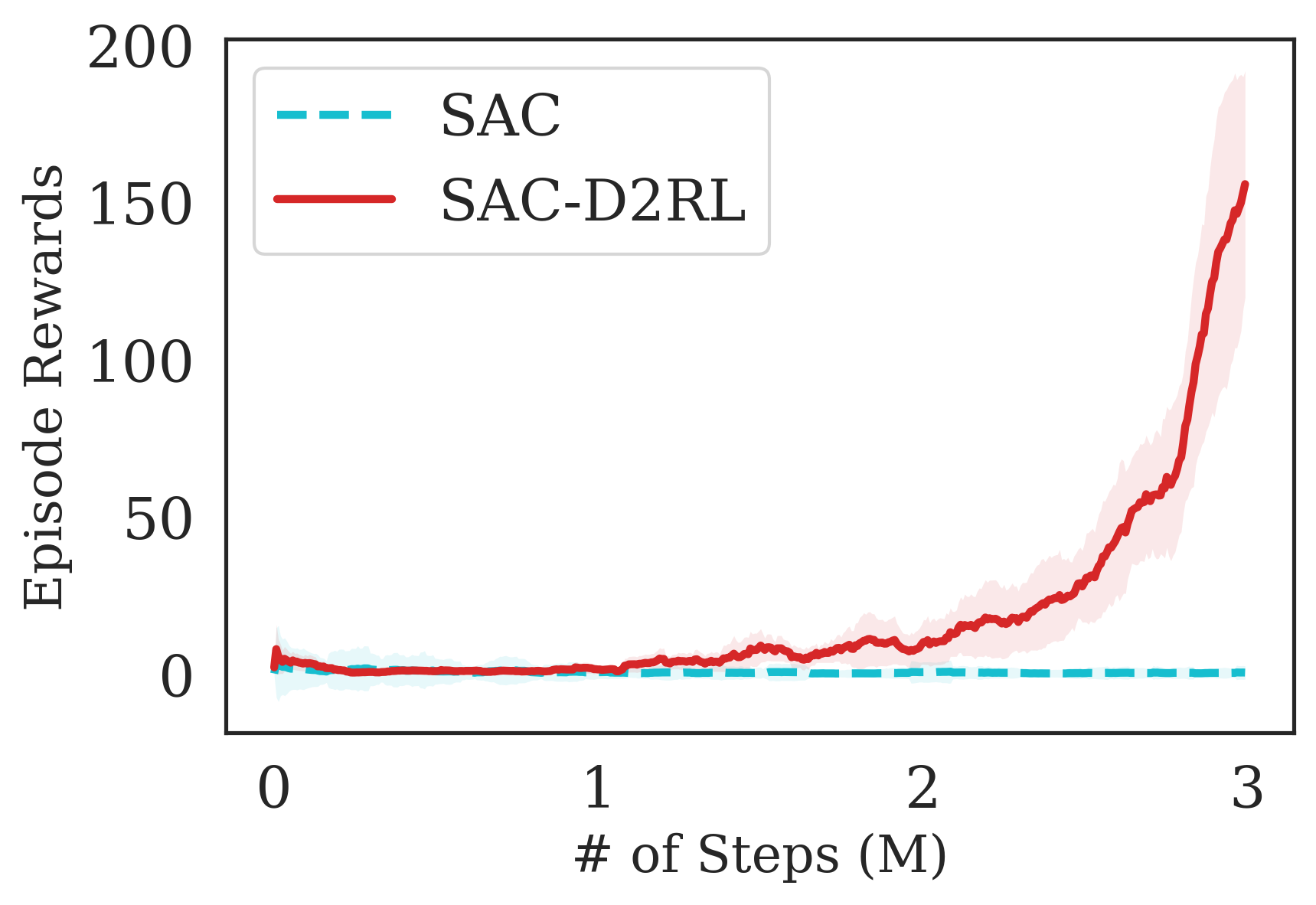}
        \caption{Jaco Reach SAC}
        \label{fig:site_sac}
    \end{subfigure}% \\
    \begin{subfigure}[b]{0.32\textwidth}
        \centering
        \includegraphics[width=\textwidth]{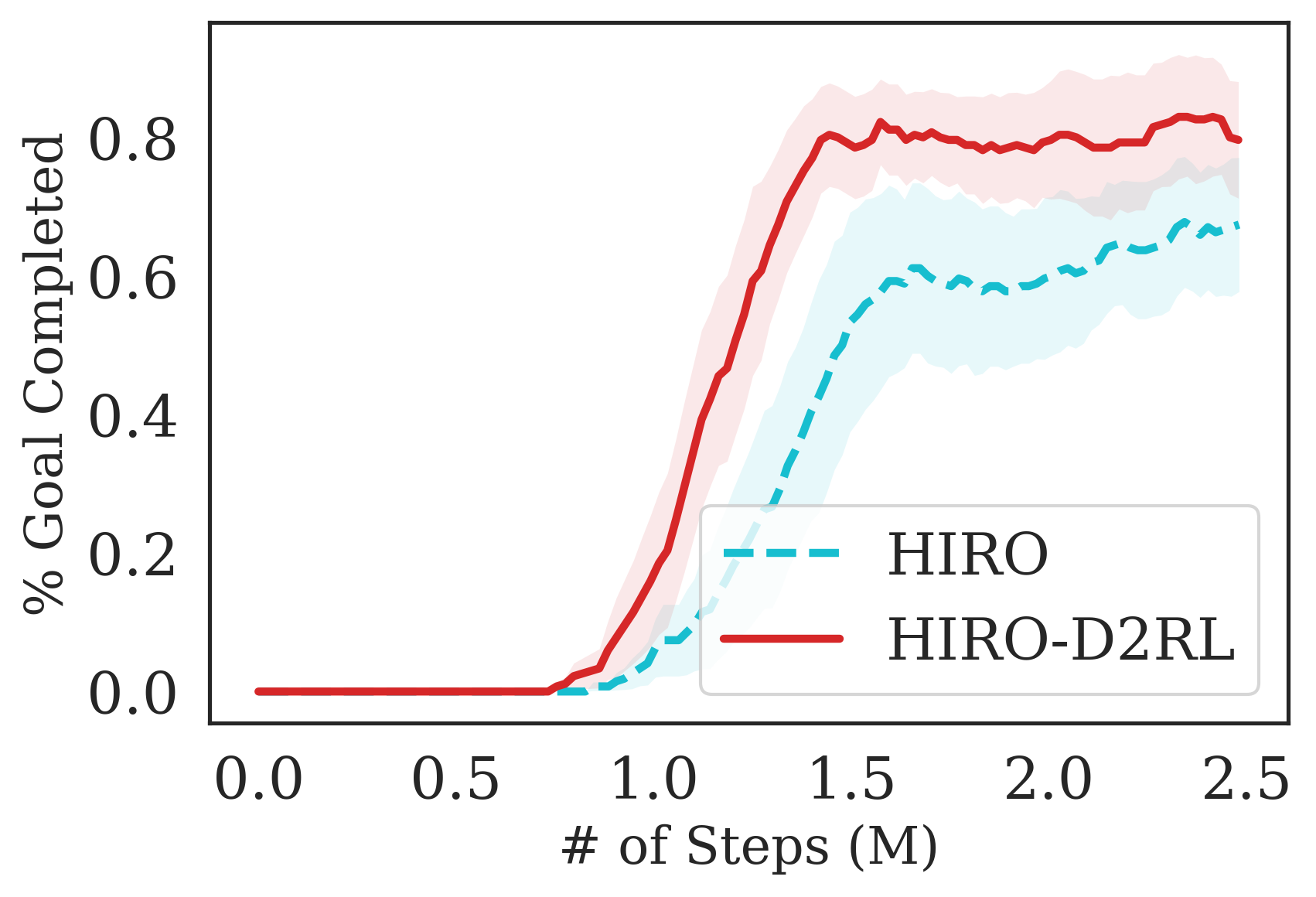}
        \caption{AntMaze HIRO}
        \label{fig:maze_hiro}
    \end{subfigure}
    \begin{subfigure}[b]{0.32\textwidth}
        \centering
        \includegraphics[width=\textwidth]{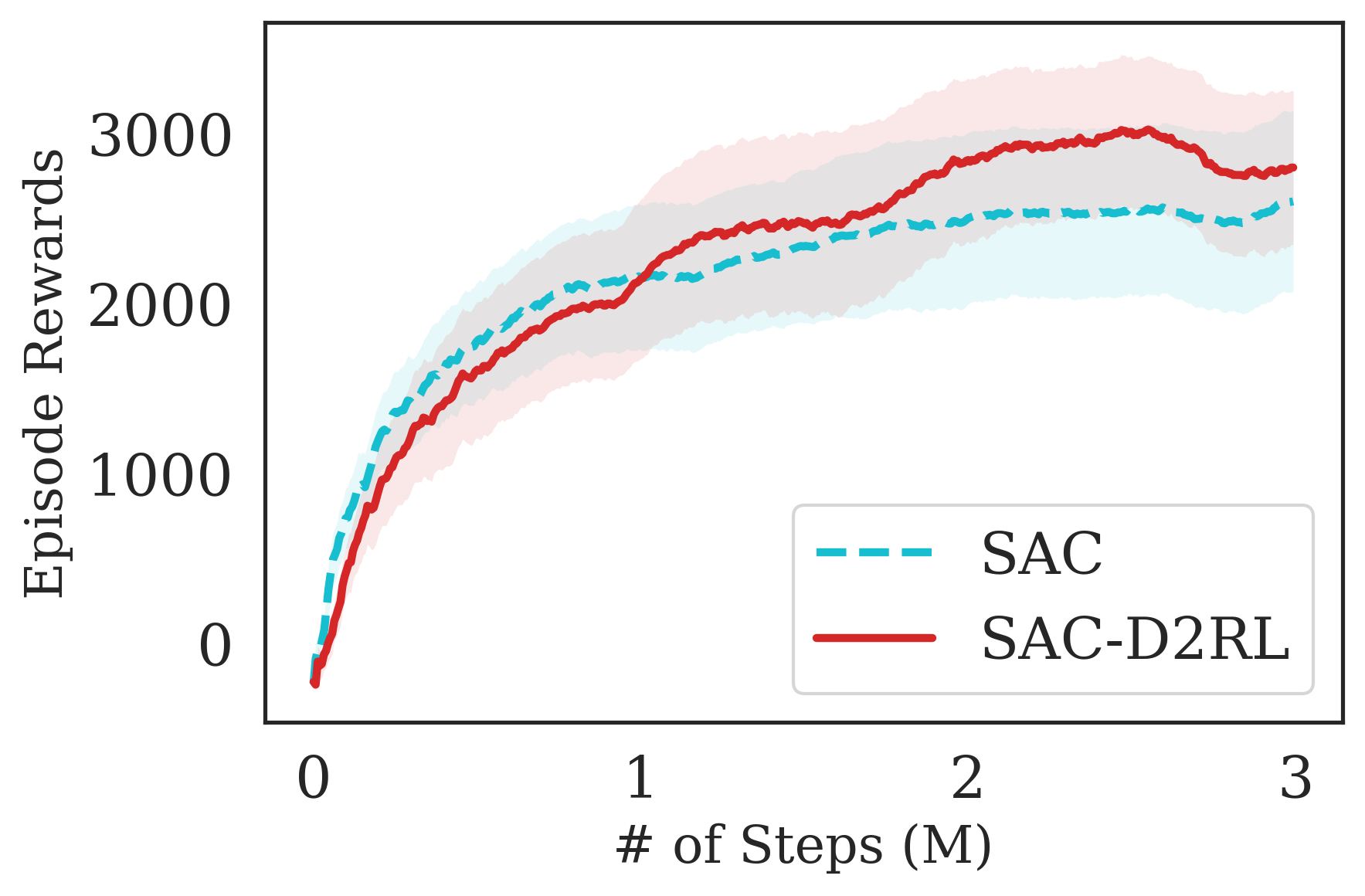}
        \caption{Baxter Block Join+Lift}
        \label{fig:furniture}
    \end{subfigure}
    %  \begin{subfigure}[b]{0.3\textwidth}
    %           \centering
    % \includegraphics[width=\textwidth]{plots/walker_td3.png}
    % \caption{Gym Walker TD3}
    % \label{fig:obstacle}
    %     \end{subfigure}
    %      \begin{subfigure}[b]{0.2\textwidth}
    %           \centering
    % \includegraphics[width=\textwidth]{pushnreach.png}
    % \caption{Sawyer Push and Reach (Image)}
    % \label{fig:pushreach}
    %     \end{subfigure}
    %\quad
     \caption{\textbf{Challenging selected manipulation and locomotion environments.} Comparison of the proposed variation \algoName and the baselines on a suite of challenging manipulation and locomotion environments. We apply the \algoName modification to the SAC~\citep{sac}, HER~\citep{her}, and HIRO~\citep{hiro} algorithms and compare relative performance in terms of average episodic rewards with respect to the baselines. The task complexity increases from Fetch Reach to Fetch Slide. Jaco Reach is challenging due to high-dimensional torque controller action space, AntMaze requires exploration to solve a temporally extended problem, and Furniture BlockJoin requires solving two tasks- join and lift sequentially. The error bars are with respect to 5 random seeds. Some additional results on the Fetch envs are in the Appendix.}
    %  \Animesh{confirm if the x-axis in fetch reach and fetch slide is not 100K, if yes change it to Millions for consistency} \Sam{yes its k not M, looked too ugly to have 0.001 M. also since this is now going to iclr and not corlm can we remove baxter block Join+lift?}}  
 \label{fig:complex}
 \vspace*{-0.5cm}
\end{figure*}

\noindent\textbf{\algoName is more sample efficient compared to baseline state-of-the-art algorithms on both image-based and state-based DM Control Suite \textit{benchmark} environments.} We compare the proposed \algoName variant with SAC and state-of-the-art pixel-based CURL algorithms on the benchmark environments of DeepMind Control Suite~\citep{dmcontrol}. For CURL, and CURL-\algoName, we train using only pixel-based observations from the environment. For SAC and SAC-\algoName, we train using proprioceptive features from the environment. The action spaces are the same in both the cases, pixels and state features based observations. The environments we consider are part of the \textit{benchamrk suite}, and include Finger Spin, Cartpole Swing, Reacher Easy, Cheetah Run, Walker Walk, Ball in Cup Catch. Additional details are in the Appendix.

In Table~\ref{tab:DML_resutls}, we tabulate results for all the algorithms after 100K environment interactions, and after 500K environment interactions. To report the results of the baseline, we simply use the results as reported in the original paper~\citep{curl}. From this Table, we observe that the \algoName variant performs better than the baselines for both 100K and 500K environment interactions, and the performance gains are especially significant in the 100K step scores. This indicates that \algoName is significantly more sample-efficient than the baseline.

\noindent\textbf{\algoName performs significantly better in challenging environments with various modalities of noise, system delays, physical perturbations and dummy-dimensions \citep{empirical_rwrl}.}
\citet{rwrl} propose a set of challenges in the DM Control Suite environments that are more ``realistic'' by introducing the aforementioned problems. 
We train a TD3 agent \cite{td3} from states, on the ``easy'' and ``medium'' challenges for the walker-walk, and cartpole-swingup environments with and without \algoName.
We present the results in  Table \ref{tab:RWRLresults}.
We see how the baseline TD3 agent gets significantly worse in the ``medium'' challenge compared to the ``easy'' version of the same environment. 
The agent trained with TD3-\algoName significantly outperforms the baseline TD3 agent on 3 of the 4 challenges, and the drop between the ``easy'' and ``medium'' challenges is significantly less severe, compared to the baseline.
This experiment demonstrates how by using \algoName, we are able to get significantly better performance on environments which have been constructed to be more realistic by adding difficult problems that the agent must learn to reason with. 
The increased robustness to such problems further validates the general utility of \algoName in many different circumstances.

\textcolor{black}{\textbf{Verifying the alleviation of implicit underparameterization with \algoName.} In Table~\ref{tab:rank}, we compare effective ranks of the feature matrices for \algoName and a normal MLP, with the TD3 algorithm after 1M interactions, corresponding to Fig.~\ref{fig:sim_resultsgym}. We observe that the effective ranks for \algoName are higher across environments. For this computation, we use the same formula of $srank_\delta(\Phi)$ from~\citep{aviral}, where $\Phi$ is the learned weight matrix of the penultimate layer of the nwtwork (i.e. the feature matrix).~\citet{aviral} showed that lower effective rank of the feature matrix correlates negatively with performance, and we believe this might help explain some of the empirical observations of better performance of \algoName. Using skip connections as in \algoName leads to higher effective rank feature matrices compared to standard MLPs used in deep RL, as can be seen from Table~\ref{tab:rank}.  }

\begin{table*}[t]
\color{black}
 \small
   \setlength\tabcolsep{1.6pt}
   \centering
%   \begin{minipage}[c]{0.65\textwidth}
%   \resizebox{0.95\linewidth}{!}{
  \centering
   \begin{tabular}{l | cc|cc| cc| cc}
     \toprule
      & \multicolumn{2}{c}{\textbf{Walker2d-v2}} & \multicolumn{2}{c}{\textbf{Ant-v2}} & \multicolumn{2}{c}{\textbf{Hopper-v2}} & \multicolumn{2}{c}{\textbf{Cheetah-v2}}\\
    \midrule
    \textbf{1M Steps}  & \textbf{TD3} & \textbf{TD3-\algoName} & \textbf{TD3} & \textbf{TD3-\algoName}& \textbf{TD3} & \textbf{TD3-\algoName}& \textbf{TD3} & \textbf{TD3-\algoName}\\
    \midrule
    Policy &  153 & \textbf{161}  &165 & \textbf{178}  &  145 & \textbf{159} &  157 & \textbf{161}\\
    \rowcolor[HTML]{EFEFEF}
    Q-network  &  138 & \textbf{165}  &153 & \textbf{180}  &  123 & \textbf{157} &  143 & \textbf{164} \\
       \bottomrule
    \end{tabular}
    % }
    %   \end{minipage}
    %   \hspace*{-0.3cm}
    % \begin{minipage}[c]{0.33\textwidth}
    \caption{\textbf{OpenAI Gym benchmark environments with TD3.} Comparison of the proposed variation \algoName and the baselines on a suite of OpenAI-Gym environments. We tabulate effective ranks of the feature matrices (learned weight matrices of the penultimate layer of the networks) for \algoName and a normal MLP, with the TD3 algorithm after 1M interactions, corresponding to Fig.~\ref{fig:sim_resultsgym}. The value is the average over 5 random seeds rounded off to the nearest integer. A loss in rank of the matrix of weights of the penultimate layers (corresponding to learning features) leads to a loss in effectively expressivity of the network, and has been shown to correlate with poor performance~\citep{aviral}. Higher is better.}
    \label{tab:rank}
    % \end{minipage}
 \end{table*}

\noindent\textbf{The sample efficiency and asymptotic performance of \algoName scale to complex robotic manipulation and locomotion environments.} Additionally, we consider some challenging manipulation and locomotion environments with different robots, the details of which are discussed below:

% \begin{itemize}
%     \item \textit{\textbf{Fetch}}-\{Reach, Pick, Push, Slide\}: There are four environments, where a Fetch robot is tasked with solving the tasks of reaching a goal location, picking an object and placing it at a goal location, pushing a puck to a goal location, and sliding a puck to a goal location. The the Fetch-Slide environment, it is ensured that sliding occurs instead of pusing because the goal location is beyong the end-effector's reach. The observations to the agent consist of proprioceptive state features and the action space is the (x,y,z) position of the end-effector and the distance between the grippers. 
%     \item \textit{\textbf{Jacko}}-Reach: A Jacko robot arm with a three finger gripper is tasked with reaching a location location indicated by a red brick. The observations to the agent consist only of proprioceptive state features and the arm is joint torque controlled.
%     \item \textit{\textbf{Ant}}-Maze: An Ant robot is tasked with navigating a U-shaped maze. The robot is joint torque controlled. 
%     \item \textit{\textbf{Baxter}}-Block Join and Lift: One arm of a Baxter robot with two fingers must be controlled to grasp a block, join it to another block and lift the combination above a certain goal height. The observations to the agent consist of proprioceptive state features and the action space is the (x,y,z) position of the end-effector and the distance between the grippers. 
% \end{itemize}

\noindent\textit{Fetch-\{Reach, Pick, Push, Slide\}}: There are four environments, where a Fetch robot is tasked with solving the tasks of reaching a goal location, picking an object and placing it at a goal location, pushing a puck to a goal location, and sliding a puck to a goal location. The the Fetch-Slide environment, it is ensured that sliding occurs instead of pusing because the goal location is beyong the end-effector's reach. The observations to the agent consist of proprioceptive state features and the action space is the (x,y,z) position of the end-effector and the distance between the grippers. 

\noindent\textit{Jaco-Reach}: A Jaco robot arm with a three finger gripper is tasked with reaching a location location indicated by a red brick. The observations to the agent consist only of proprioceptive state features and the arm is joint torque controlled.

\noindent\textit{Ant-Maze}: An Ant robot with four legs is tasked with navigating a U-shaped maze while being joint torque controlled. This is a challenging locomotion environment with a temporally-extended task, that requires the agent to move around the maze to reach the goal. 

\noindent\textit{Baxter-Block Join and Lift}: One arm of a Baxter robot with two fingers must be controlled to grasp a block, join it to another block and lift the combination above a certain goal height. The observations to the agent consist of proprioceptive state features and the action space is the (x,y,z) position of the end-effector and the distance between the grippers. 

For the \textit{\textbf{Fetch}}-\{Reach, Pick, Push, Slide\} environments, we consider the HER~\citep{her} algorithm (with DDPG~\citep{ddpg}) trained with sparse rewards that was shown to achieve state-of-the-art results in these environments. The plots for \textbf{\textit{Fetch}}- Pick and \textbf{\textit{Fetch}}-Push are in the Appendix, sue to space constraint here. In addition, we also show results with SAC on \textbf{\textit{Fetch}}-Reach and \textbf{\textit{Fetch}}-Slide trained using a SAC agent. 
For \textit{\textbf{Ant}}-Maze, we consider the hierarchical RL algorithm HIRO~\citep{hiro} that was shown to be successful in this very long horizon task. 
For \textit{\textbf{Jaco}}-Reach and \textit{\textbf{Baxter}}-Block Join and Lift, we consider the default SAC algorithm released with the environment codebase \url{https://github.com/clvrai/furniture}

The results are summarized in Fig.~\ref{fig:complex}, where we see that the proposed \algoName modification converges to higher episodic rewards and converges significantly faster in most environments.
By performing a wide range of experiments on challenging robotics environments, we further notice significantly better sample efficiency on all environments which suggests the wide generality and applicability of \algoName.
Interestingly, we also observe in \ref{fig:site_sac} that SAC is unable to train the agent to perform the Jaco-Reach task in 3M environment steps, while SAC trained with \algoName policy and $Q$-networks is able to succesfully train an agent and starts outperforming the SAC baseline as early as 1M environment steps. 
This shows how crucial parameterization is in some environments as a simple 2-layer MLPs may not be sufficiently expressive or optimization using deeper network architectures may be necessary to solve such environments.

\subsection{Comparisons with ResNet}

\textcolor{black}{
Table \ref{tab:DML_resutls} tabulates experiments with a ResNet-like MLP which 
utilizes residual connections instead of dense connections \cite{resnet}.
We experiment with images using a base CURL agent \cite{curl}, and directly from proprioceptive
features using a base SAC agent \cite{sac} on the DM Control Suite.
We see that using \algoName clearly outperforms the ResNet agent on most tasks from 
images and state features.
Residual connections add the features of a previous layer to the current layer 
instead of concatenating them (as done in \algoName). 
The same hyperparameters are used as \algoName for all experiments.
}
\subsection{Ablation Studies}
\vspace*{-0.2cm}
\label{sec:ablation}

\begin{figure*}[h!]
\centering
  \begin{minipage}[c]{0.58\textwidth}
  \resizebox{0.98\linewidth}{!}{
  \centering
    \begin{subfigure}[b]{0.5\textwidth}
        \centering
        \includegraphics[width=\textwidth]{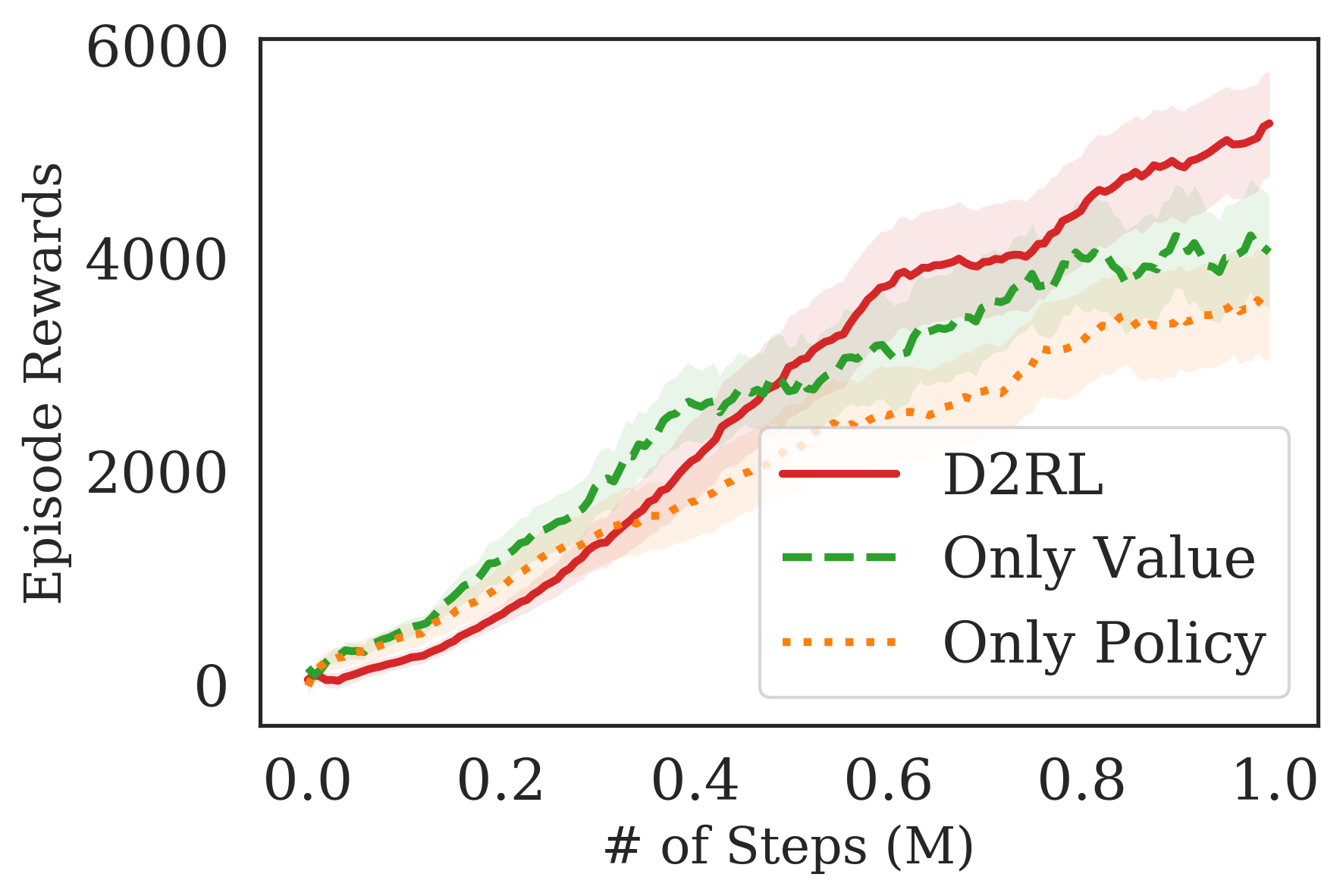}
        \caption{Importance of \algoName}
        \label{fig:ablation_imp}
    \end{subfigure}% \\
    \begin{subfigure}[b]{0.5\textwidth}
        \centering
        \includegraphics[width=\textwidth]{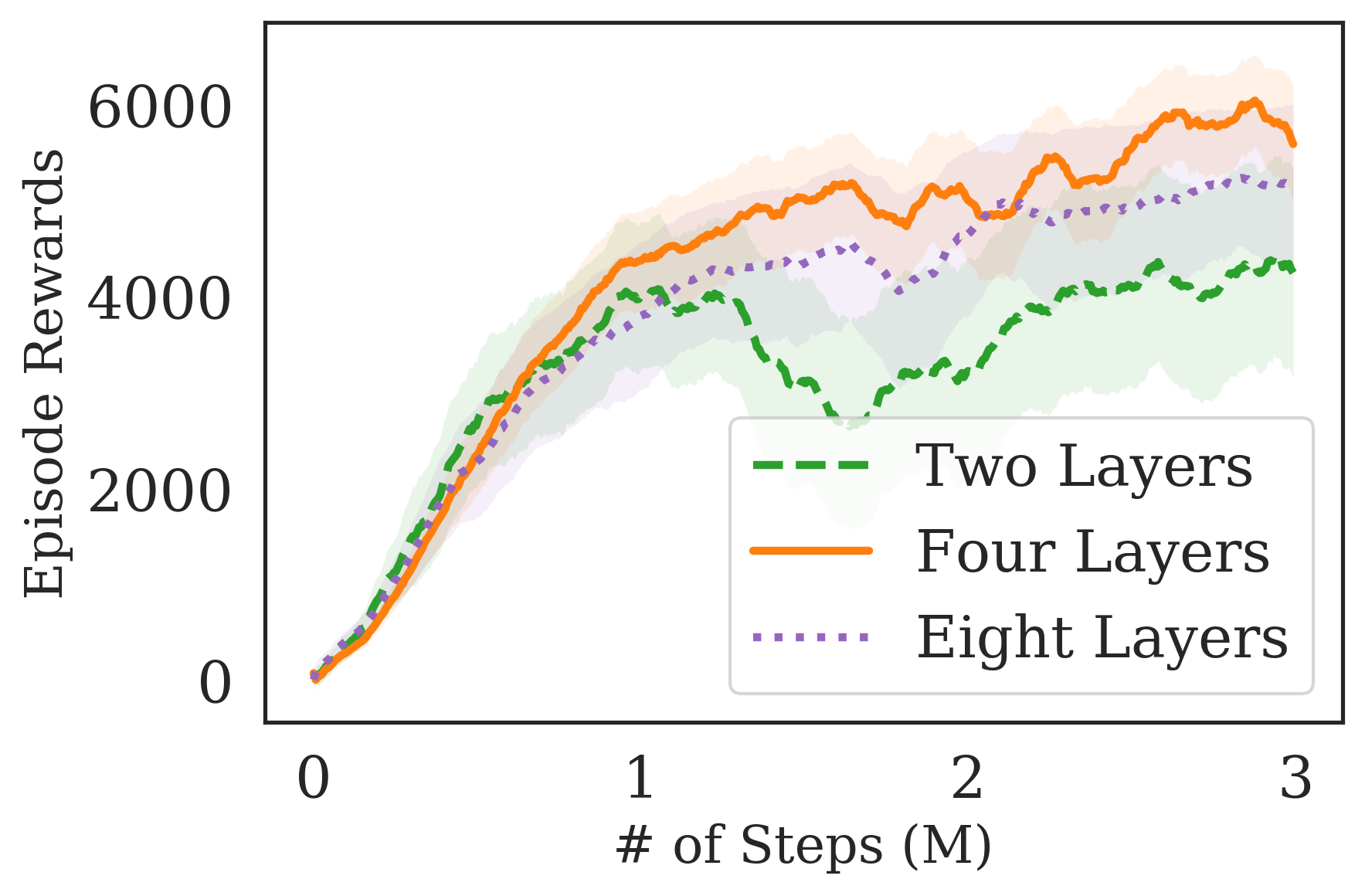}
        \caption{Number of layers in \algoName}
        \label{fig:ablation_layer}
    \end{subfigure}
    }
\end{minipage}    
  \begin{minipage}[c]{0.41\textwidth}
    \caption{Ablation studies with a SAC agent on the Ant-v2 env in the Open AI Gym suite. It is evident that the \algoName policy architecture applied to both the policy and Q-value networks achieves higher rewards than being applied to either just the policy or just the value network. Also, in (b) deeper \algoName networks perform better, in contrast to vanilla MLP networks in Fig.~\ref{fig:num_layers}
    % \aravind{could you add like a conclusive message to the caption, and maybe make the figure bigger? y-axis/x-axis labels are tiny.}\Homanga{Done!}
    }
 \label{fig:ablation}
 \end{minipage}
 \vspace*{-0.2cm}
\end{figure*}

In this section, we look to answer the various components of using \algoName.
We first analyze how the agent performs when only the policy or the value functions are parameterized as \algoName, while the other one is a vanilla 2-layer MLP.
The results for training an SAC agent on Ant-v2 are present in Fig. \ref{fig:ablation_imp}, where we see that parameterizing both the networks as \algoName significantly outperforms when only one of the two use \algoName.
However, one noteworthy observation can be made that when only the value functions are parameterized using \algoName, the agent significantly outperforms the variant where only the policy is parameterized using a \algoName.
This suggests that it may be more important to parameterize the value function, but more research is required to give a more conclusive statement.

Similarly we train the same agent but instead vary the number of layers used while parameterizing the policies and value functions using \algoName. The results in Fig. \ref{fig:ablation_layer} show that even when 8 layer \algoName is used, the results are only moderately worse that when using 4 layers, even though it has twice the depth and therefore twice as many parameters. These results are notably different from the results in Fig. \ref{fig:num_layers}, where as we increase the number of layers for vanilla MLPs to be greater than 2, we see a worsening results. The difference suggests that by using \algoName we are able to circumvent the issue of DPI that may hinder the performance for vanilla MLPs, as we postulated.

\vspace*{-0.1cm}
\section{Discussion, Future Work, and Conclusion}
\vspace*{-0.1cm}
In this paper, we investigated the effect of building better inductive biases into the architectures of the function approximators in deep reinforcement learning. 
We first looked into the effect of varying the number of layers to parameterize policies and value functions, and how the performance deteriorates as the number of layers increase.
To overcome this problem, we proposed a generally applicable solution that significantly improves the sample efficiency of the state-of-the-art DRL baselines over a variety of manipulation, and locomotion environments with different robots, from both states and images.
Studying the effect of network architectures has been long explored in computer vision and deep learning, and its benefits on performance have been well established. The effect of the network architectures, however, have not yet been studied in DRL and robotics. Improving the network architectures for a variety of popular standard actor-critic algorithms demonstrates the importance of building better inductive biases in the network paramterization such that we can improve the performance for the otherwise identical algorithms. 
In future work, we are interested in building better network architectures and further improving the underlying algorithms for robotic learning.

\section*{Acknowledgement}
We thank Vector Institute Toronto for compute support, and Mayank Mittal, Irene Zhang, Alexandra Volokhova, Kevin Xie, and other members of the UofT CS Robotics group for helpful discussions and feedback on the draft. We thank Denis Yarats (NYU) for feedback on a previous version of the paper and for pointing out the findings about network architecture in SAC-AE. We thank Richard Song (Google) for providing insights on using wider networks in RL and the implications this could have in terms of generalization. We are grateful to Kei Ohta (Mitsubishi) for independently reproducing a Tensorflow 2 version of D2RL. HB thanks Aviral Kumar (UC Berkeley) for suggesting the connection D2RL might have in alleviating implicit under-parameterization in Deep Q learning. 

\bibliography{iclr2021_conference}  % .bib
\bibliographystyle{iclr2021_conference}

\newpage

\appendix
% \section*{Note about Real Robot Experiments}
% In the results reported in the paper, we performed extensive evaluation across different robotic control environments, different robot dynamics, different tasks, and different controllers. In particular, we considered OpenAI Gym MuJoCo environments (locomotion tasks involving agents like Half-Cheetah, Hopper, Walker, Humanoid, and Ant), Fetch robot environments (reaching, pushing, sliding, pick and place; End-effector position controlled robot), Jaco robot environment (reaching; joint torque controlled robot), Baxter robot environment (joining and lifting two blocks; End-effector position controlled robot), and an Ant maze navigation environment (joint torque controlled robot). In addition, we reported results using both state-based proprioceptive features and images (in the CURL-based results in Table 1) as observations. We believe that the diversity of environments considered, different controllers, and different forms of supervision (states and images) demonstrate the potential applicability of our approach to real robots too.   

%\newpage
\section{Pytorch Code}
\label{pytorch_code}

\begin{lstlisting}[language=Python,label={lst:pytorch_alg}]
import torch
import torch.nn as nn
import torch.nn.functional as F

LOG_SIG_MAX = 2
LOG_SIG_MIN = -5

class Policy(nn.Module):
    def __init__(self, num_inputs, num_actions, hidden_dim, action_space=None):
        super(Policy, self).__init__()
        in_dim = hidden_dim+num_inputs
        self.linear1 = nn.Linear(num_inputs, hidden_dim)
        self.linear2 = nn.Linear(in_dim, hidden_dim)
        self.linear3 = nn.Linear(in_dim, hidden_dim)
        self.linear4 = nn.Linear(in_dim, hidden_dim)
        self.mean_linear = nn.Linear(hidden_dim, num_actions)
        self.log_std_linear = nn.Linear(hidden_dim, num_actions)

    def forward(self, state):
        x = F.relu(self.linear1(state))
        x = torch.cat([x, state], dim=1)
        x = F.relu(self.linear2(x))
        x = torch.cat([x, state], dim=1)
        x = F.relu(self.linear3(x))
        x = torch.cat([x, state], dim=1)
        x = F.relu(self.linear4(x))

        mean = self.mean_linear(x)
        log_std = self.log_std_linear(x)
        log_std = torch.clamp(log_std, min=LOG_SIG_MIN, max=LOG_SIG_MAX)
        return mean, log_std
        
class QNetwork(nn.Module):
    def __init__(self, num_inputs, num_actions, hidden_dim, num_layers):
        super(QNetwork, self).__init__()

        in_dim = num_inputs+num_actions+hidden_dim
        self.l1_1 = nn.Linear(num_inputs+num_actions, hidden_dim)
        self.l1_2 = nn.Linear(in_dim, hidden_dim)
        self.l1_3 = nn.Linear(in_dim, hidden_dim)
        self.l1_4 = nn.Linear(in_dim, hidden_dim)

        self.out1 = nn.Linear(hidden_dim, 1)

    def forward(self, state, action):
        xu = torch.cat([state, action], dim=1)
        x1 = F.relu(self.l1_1(xu))
        x1 = torch.cat([x1, xu], dim=1)
        x1 = F.relu(self.l1_2(x1))
        x1 = torch.cat([x1, xu], dim=1)
        x1 = F.relu(self.l1_3(x1))
        x1 = torch.cat([x1, xu], dim=1)
        x1 = F.relu(self.l1_4(x1))
        
        x1 = self.out1(x1)
        return x1
\end{lstlisting}

PyTorch code for a stochastic SAC policy and $Q$-network~\citep{sac}.
The code provided can simply replace the policy and $Q$-network for any current SAC implementation, or be adopted for other actor-critic algorithms such as TD3~\citep{td3} or DDPG~\citep{ddpg}.

\section{Additional Experiments}
\label{additional_exps}

\begin{figure*}[h!]
\centering
      \begin{subfigure}[b]{0.32\textwidth}
        \centering
        \includegraphics[width=\textwidth]{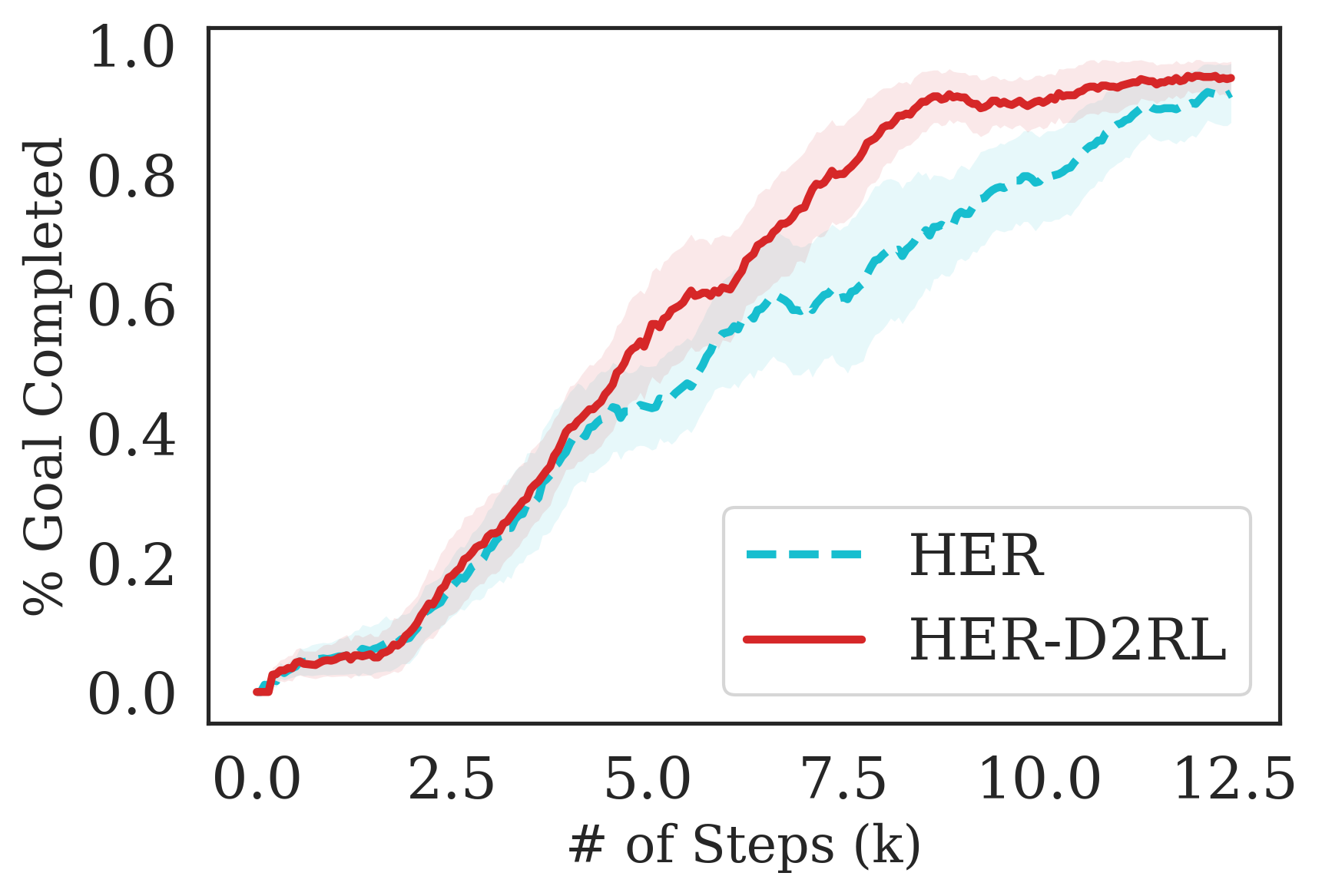}
        \caption{Fetch Pick HER}
        \label{fig:reach_sac}
    \end{subfigure}
        \begin{subfigure}[b]{0.32\textwidth}
        \centering
        \includegraphics[width=\textwidth]{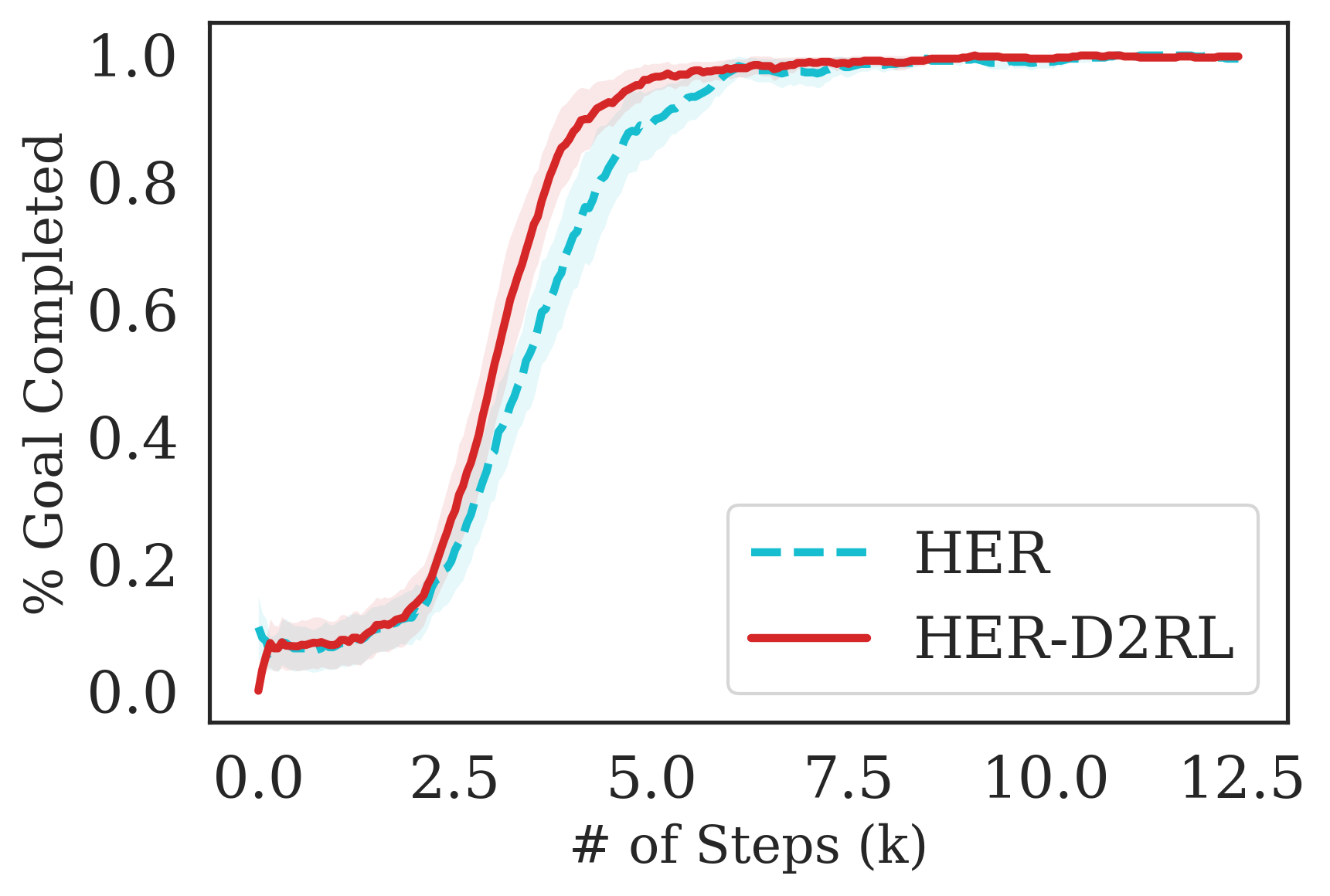}
        \caption{Fetch Push HER}
        \label{fig:slide_sac}
    \end{subfigure}
      \begin{subfigure}[b]{0.32\textwidth}
        \centering
        \includegraphics[width=\textwidth]{plots/slide_her.png}
        \caption{Fetch Slide HER}
        \label{fig:narrow}
    \end{subfigure}
    \caption{Complete set of experiments with HER~\citep{her} and DDPG~\citep{ddpg} on the Fetch robot. Using \algoName continues to outperform the baseline on the three environments considered.}
 \label{fig:complex}
 \vspace*{-0.3cm}
\end{figure*}
  
% \begin{figure*}[h!]
% \centering
%       \begin{subfigure}[b]{0.32\textwidth}
%         \centering
%         \includegraphics[width=\textwidth]{plots/humanoid_sac.png}
%         \caption{Gym Humanoid SAC}
%         \label{fig:reach_sac}
%     \end{subfigure}
%         \caption{\textbf{OpenAI Gym benchmark environments.} Comparison of the proposed variation \algoName and the baseline on Humanoid-v2 (continuation of Fig. 4). The error bars are with respect to 5 random seeds.}  
%  \label{fig:humanoid}
%  \vspace*{-0.3cm}
% \end{figure*}

\section{Hyperparameters and Environment details}
\label{hyperparams}

\begin{table}[h!]
\caption{Hyperparameters used for SAC in all the experiments. The hyperparameter values are kept the same across SAC and SAC-\algoName baselines in all the environments.}

\label{table:hyperparameters}
\vskip 0.15in
\begin{center}
\begin{small}
\begin{tabular}{ll}
\toprule
\textbf{Hyperparameter} & \textbf{Value}  \\
\midrule
Hidden units (MLP)    & $256$ for Gym  \\ 
 & $1024$ for DM Control \citep{sac_ae}\\
Evaluation episodes    & $10$  \\ 
Optimizer    & Adam  \\ 
Learning rate $(\pi_\psi, Q_\phi)$   & $3e-4$ \\ 
Learning rate ($\alpha$) & $1e-4$ \\
MiniBatch Size    & $256$ \\ 
Non-linearity & ReLU \\
Discount $\gamma$ & $.99$ \\
Initial temperature & $0.1$ \\
\bottomrule
\end{tabular}
\end{small}
\end{center}
\vskip -0.1in
\end{table}

\begin{table}[h]
\caption{Hyperparameters used for DMControl CURL experiments. The hyperparameter values are kept the same across CURL, CURL-\algoName, and CURL-ResNet baselines.}

\label{table:hyperparameters}
\vskip 0.15in
\begin{center}
\begin{small}
\begin{tabular}{ll}
\toprule
\textbf{Hyperparameter} & \textbf{Value}  \\
\midrule
Random crop    & True  \\ 
Observation rendering    & $(100,100)$  \\ 
Observation downsampling    & $(84,84)$  \\ 
Replay buffer size    & $100000$ \\ 
Initial steps    & $1000$  \\ 
Stacked frames    & $3$  \\ 
Action repeat    & $2$ finger, spin; walker, walk\\
 & $8$ cartpole, swingup \\
 & $4$ otherwise  \\
Hidden units (MLP)    & $1024$ \citep{sac_ae}  \\ 
Evaluation episodes    & $10$  \\ 
Optimizer    & Adam  \\ 
$(\beta_1,\beta_2) \rightarrow (f_\theta, \pi_\psi, Q_\phi)$   & $(.9,.999)$  \\
$(\beta_1,\beta_2) \rightarrow (\alpha)$   & $(.5,.999)$  \\
Learning rate $(f_\theta, \pi_\psi, Q_\phi)$     & $3e-4$ \\
Learning rate ($\alpha$) & $1e-4$ \\

Batch Size    & $512$ (cheetah), 128 (rest)  \\ 
$Q$ function EMA $\tau$ & $0.01$ \\
Critic target update freq & $2$ \\
Convolutional layers & $4$ \\
Number of filters & $32$ \\
Non-linearity & ReLU \\
Encoder EMA $\tau$ & $0.05$ \\
Latent dimension & $50$ \\
Discount $\gamma$ & $.99$ \\
Initial temperature & $0.1$ \\

\bottomrule
\end{tabular}
\end{small}
\end{center}
\vskip -0.1in
\end{table}

\begin{figure}[h!]
    \centering
    \includegraphics[width=0.9\textwidth]{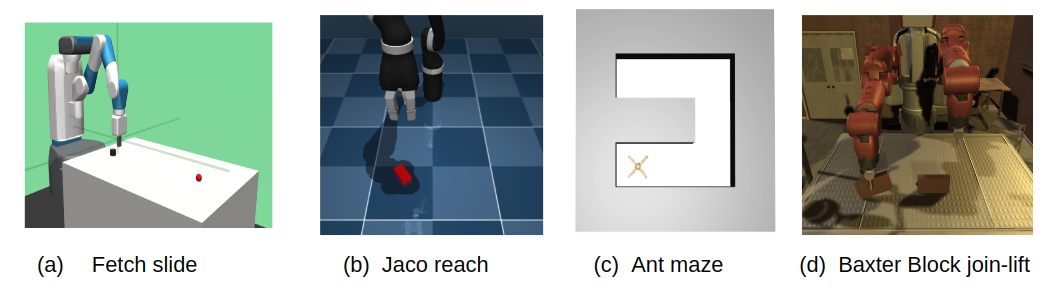}
    \caption{Illustrations of some of the challenging robotic control environments used for our experiments. In \textit{Fetch slide}, a Fetch robot arm with one finger must be controlled to slide a puck to a goal location. In \textit{Jacko reach}, a Jaco robot with a three finger gripper must be controlled to reach the red brick. In \textit{Ant maze}, an Ant with four legs must be controlled to navigate a maze. In \textit{Baxter block join-lift}, one arm of a Baxter robot with a two finger gripper must be controlled to join two blocks and lift the combination above a certain height.}
    \label{fig:envs}
    \vspace{-15pt}
\end{figure}

\begin{table}[h!]
\caption{Consolidated details of all the environments used in our experiments}
\label{table:environmentdetails}
\vskip 0.15in
\begin{center}
\begin{small}
\begin{tabular}{cccccc}
\toprule
\textbf{Environment} & \textbf{Type} & \textbf{Controller} & \textbf{Inputs} & \textbf{Action dim.} & \textbf{Input dim.}\\
\midrule
Gym Cheetah    & Locomotion & Joint torque & States & 6  & 17\\ 
Gym Hopper    & Locomotion & Joint torque & States & 3  & 11\\ 
Gym Humanoid    & Locomotion & Joint torque & States & 17  & 376\\ 
Gym Walker    & Locomotion & Joint torque & States & 6  & 17\\ 
Gym Ant    & Locomotion & Joint torque & States & 8  & 111\\ 
Finger, Spin    & Classical control & Joint torque & States/Images & 2  & 9 / 84x84x3\\ 
Cartpole, Swing    & Classical control & Joint torque & States/Images & 1  & 5 / 84x84x3\\ 
Reacher, Easy   & Classical control & Joint torque & States/Images & 2  & 6 / 84x84x3\\ 
Cheetah, Run   & Locomotion & Joint torque & States/Images & 6  & 17 / 84x84x3\\ 
Walker, Walk   & Locomotion & Joint torque & States/Images & 6  & 24 / 84x84x3\\ 
Ball in a Cup, Catch   & Manipulation & Joint torque & States/Images & 2  & 8 / 84x84x3\\ 
Fetch Reach   & Manipulation & EE position & States & 4  & 10\\ 
Fetch Pick and Place   & Manipulation & EE position & States & 4  & 25 \\ 
Fetch Push   & Manipulation & EE position & States & 4  & 25\\ 
Fetch Slide   & Manipulation & EE position & States & 3  & 25\\ 
Jaco Reach   & Manipulation & Joint torque & States & 9  & 45\\ 
Baxter JoinLift   & Manipulation & Joint torque & States & 15  & 42 \\ 
Ant Maze   & Locomotion & Joint torque & States & 8  & 30\\ 
\bottomrule
\end{tabular}
\end{small}
\end{center}
\vskip -0.1in
\end{table}

\end{document}